\documentclass[review]{elsarticle}
\usepackage{hyperref}
\usepackage{times}
\usepackage{graphicx}
\usepackage{algorithm,algorithmic}
\usepackage{amsfonts,amssymb,amsmath,amsthm} 
\usepackage{xcolor}
\usepackage{url}
\usepackage{multirow}
\usepackage{booktabs}
\usepackage{enumitem}
\usepackage{subfig}

\theoremstyle{plain}
\newtheorem{theorem}{Theorem}

\theoremstyle{definition}

\DeclareMathOperator*{\argmin}{arg\,min}

\DeclareMathOperator*{\E}{\mathbb{E}}
\newcommand{\st}{\mathrm{s.t.}}

\newcommand{\bx}{\boldsymbol{x}}
\newcommand{\bw}{\boldsymbol{w}}
\newcommand{\by}{\boldsymbol{y}}
\newcommand{\bz}{\boldsymbol{z}}

\newcommand{\bu}{\boldsymbol{u}}
\newcommand{\bv}{\boldsymbol{v}}

\newcommand{\Rd}{\mathbb{R}^d}

\newcommand{\svmrank}{SVM$^{\textrm{rank}}$}

\graphicspath{{fig/}}

\journal{Journal of \LaTeX\ Templates}

\begin{document}

\begin{frontmatter}

\title{Object Proposal with Kernelized Partial Ranking}

\author[mymainaddress,mysecondaryaddress]{Jing Wang}
\author[mysecondaryaddress]{Jie Shen}
\author[mymainaddress,mysecondaryaddress]{Ping Li}

\address[mymainaddress]{Department of Statistics and Biostatistics, Rutgers University,  Piscataway, NJ 08854, USA}
\address[mysecondaryaddress]{Department of Computer Science, Rutgers University, Piscataway, NJ 08854, USA}

\begin{abstract}
Object proposals are an ensemble of bounding boxes with high potential to contain objects. In order to determine a small set of proposals with a high recall, a common scheme is extracting multiple features followed by a ranking algorithm which however, incurs two major challenges: {\bf 1)} The ranking model often imposes pairwise constraints between each proposal, rendering the problem away from an efficient training/testing phase; {\bf 2)} Linear kernels are utilized due to the computational and memory bottleneck of training a kernelized model.

\vspace{0.05in}
In this paper, we remedy these two issues by suggesting a {\em kernelized partial ranking model}. In particular, we demonstrate that {\bf i)} our partial ranking model reduces the number of constraints from $O(n^2)$ to $O(nk)$ where $n$ is the number of all potential proposals for an image but we are only interested in top-$k$ of them that has the largest overlap with the ground truth; {\bf ii)} we permit non-linear kernels in our model which is often superior to the linear classifier in terms of accuracy. For the sake of mitigating the computational and memory issues, we introduce a consistent weighted sampling~(CWS) paradigm that approximates the non-linear kernel as well as facilitates an efficient learning. In fact, as we will show, training a linear CWS model amounts to learning a kernelized model. Extensive experiments demonstrate that equipped with the non-linear kernel and the partial ranking algorithm, recall at top-$k$ proposals can be substantially improved.
\end{abstract}


\end{frontmatter}


\section{Introduction}
Objectness is an emerging topic in the computer vision community proposed by~\cite{alexe2010object}, which aims to produce an ensemble of regions (i.e., object proposals) that have high probability to contain objects. The main advantage of object proposal is that it can dramatically reduce the search space from millions of positions, scales and aspect ratios to hundreds of suggested candidates while ensuring a high recall.  Therefore, it is an important technique for further vision tasks such as object recognition, detection and scene understanding~\cite{girshick2013rich,sermanet2013overfeat,simonyan2014very,ren2015faster}.

Since in most scenarios, object proposal actually serves as a preprocessing step, several important ingredients should be considered for a successful proposal algorithm. First, the algorithm should be fast enough. Otherwise, its superiority to the sliding window paradigm will be degraded. Second, it should produce a manageable number of proposals with a high recall.

To this end, a large body of works are devoted to effective features and fast grouping strategies. For example, in the work of~\cite{cheng2014bing}, Cheng~et al. designed a binary feature descriptor termed ``Bing'' and trained a linear model to estimate the locations of objects. Their algorithm is computationally efficient without loss of much accuracy. In~\cite{sermanet2013overfeat,erhan2014scalable,DBLP:conf/nips/SzegedyTE13}, they utilized deep convolutional networks to extract features. The deep network mainly used GPU to speed up the process. Specifically, Ren et al. introduced a region proposal network which could be combined with the Zeiler and Fergus model~\cite{zeiler2014visualizing} and the Simonyan and Zisserman model~\cite{simonyan2014very}\cite{ren2015faster}. In~\cite{rahtu2011learning,uijlings2013selective}, Uijlings~et al. started with the low level super-pixels and carefully designed some simple yet effective features that could deal with a variety of image conditions. Then proposals were generated by grouping the super-pixels according to the handcrafted features. As there is not much computational cost in the grouping process, their algorithm is efficient. Notably, their model is fully unsupervised and hence no parameter will be learned or tuned. In~\cite{carreira2010constrained,endres2010category,arbelaez2014multiscale}, various visual cues, such as segmentation and saliency were utilized to describe a candidate region. Subsequently, based on the similarity of region features, a hierarchical grouping strategy was adopted to form the final object proposals. \cite{arbelaez2014multiscale} proposed a multi-scale hierarchical segmentation and grouped multiscale regions by features of size/location, shape and contours .

\begin{figure*}[t]
	\centering
	\subfloat[Original image]{
		\includegraphics[width=0.24\textwidth]{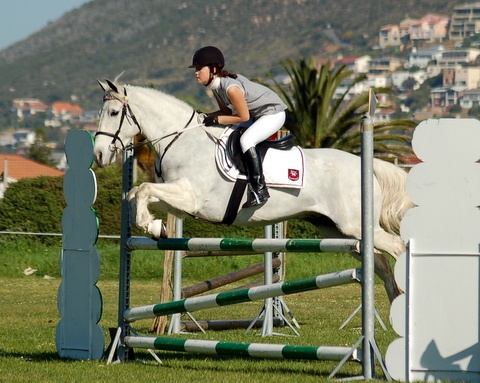}}
	\subfloat[Proposals]{
		\includegraphics[width=0.24\textwidth]{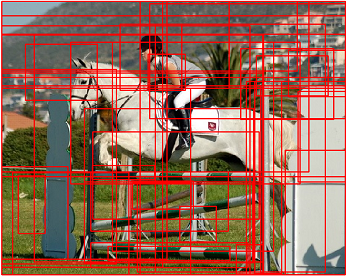}}
	\subfloat[\svmrank]{
		\includegraphics[width=0.24\textwidth]{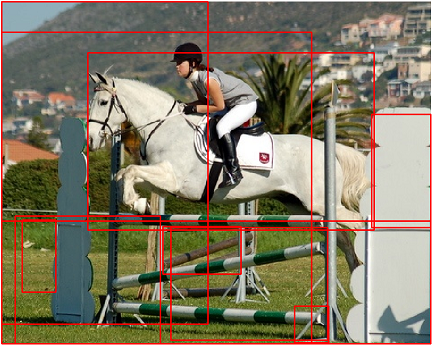}}
	\subfloat[Top-$k$ by ours]{
		\includegraphics[width=0.24\textwidth]{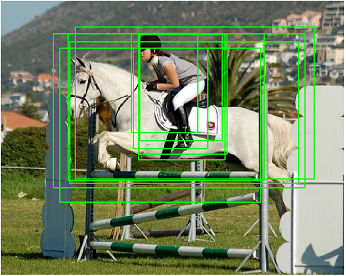}}
	\caption{\textbf{Illustration of the importance of accurate estimation for top-$k$ object proposals.} Given an image shown as (a), existing object proposal methods (e.g., Selective Search) generate a set of proposals as illustrated in (b). Typically, \emph{only} the top-$k$ candidates are used to feed further vision tasks like object detection. In (c) and (d), we visualize the top-$k$ results produced by \svmrank~ $\ $and our model respectively. Clearly, our ranking model is superior to (c) since there is fewer inaccurate proposals within the top-$k$ candidates.}
	\label{figShow}
\end{figure*}

Usually, a proposal algorithm tends to produce a large number of candidates. Hence, existing algorithms always provide a confidence score for each candidate which indicates the probability of containing an object. Commonly used schemes for the objectness scoring are summarized in~\cite{zitnick2014edge,Hosang2015pami}. Among them, the large margin based \svmrank, or its variant is a popular solution~\cite{endres2010category,rahtu2011learning,cheng2014bing,arbelaez2014multiscale,zhang2016object}. Given all the candidates of an image, \svmrank$\ $considers the pair-wise ranks as constraints. However, imposing such full rankings for each candidate is possibly not necessary, and sometimes over constrained. To see this, consider the case that we have two candidates with  Intersection Over Union (IoU) 0.01 and 0.001. Actually, they both can be treated as incorrect proposals. In this case, constraining the first candidate to have a higher rank than the other does not help much for the model construction. As we only care about the top-$k$ candidates, a full ranking algorithm such as \svmrank$\ $ is not suitable for object proposals location. In Figure~\ref{figShow}, we give an example showing that an accurate prediction for the top-$k$ candidates is more important than obtaining the rank for all candidates.

Related to the ranking algorithm, previous works usually devise hand-crafted features and feed them to a linear predictor. Yet, as has been known, non-linear kernels are often superior to the linear one in terms of prediction accuracy. One possible shortcoming of non-linear kernel is the memory and computation bottleneck. Fortunately, recent progress demonstrates that a class of popular kernels can be approximated by linear functions, such as shift-invariant kernels~\cite{rahimi2007random} and generalized min-max (GMM) kernels~\cite{Report:Li_GMM16}.

\begin{figure*}[t]
	\centering
	\includegraphics[width=1\linewidth]{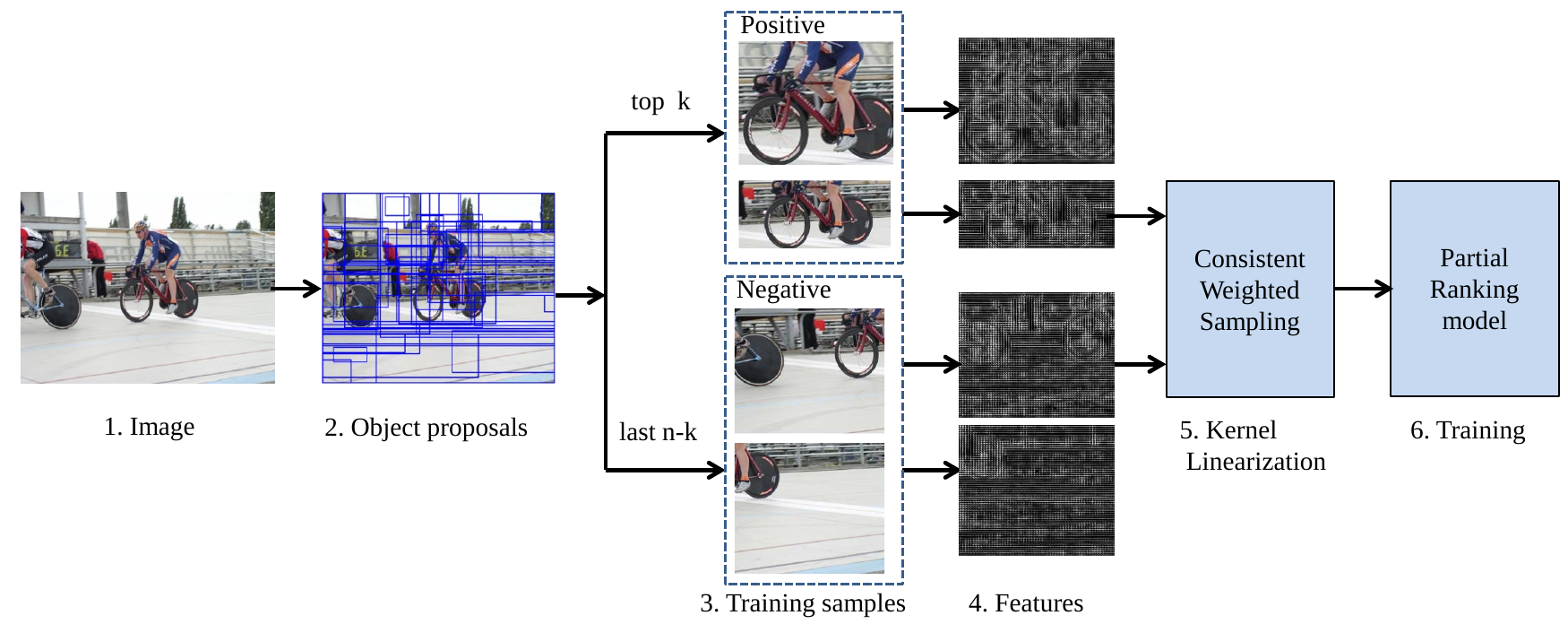}
	
	\caption{Overview of the learning procedure. Our system (1) takes an input image, (2) obtains proposals which are produced by some previous proposal algorithm such as Selective Search, (3) splits the set of candidates into the top-$k$ subset and the last $n-k$ subset according to the IoU to the ground truth, (4) computes features for each proposal followed by (5) consistent weighted sampling, and then (6) learns the partial ranking model based on the output of CWS.
	}
	\label{fig:overview}
\end{figure*}

In this paper, henceforth, we propose a new partial ranking algorithm with support of non-linear kernel. The overview of the procedure is illustrated in Figure~\ref{fig:overview}. Given the ground truth and an ensemble of object candidates which are produced by existing methods, we compute the IoU for each candidate and then split these potential objects into two subsets, one of which consists of the top-$k$ candidates and the remaining forms another group. The feature used here is the popular HOG~\cite{dalal2005histograms}, which will be described in Section~\ref{subsec:feature}. Yet, one can also replace it with other popular descriptors, such as SIFT or CNN features. Then we perform (0-bit) consistent weighted sampling~(CWS)~\cite{cws,Proc:Ioffe_ICDM10,min-max,Report:Li_GMM16} on the features followed by learning our partial ranking model. In this way, learning a ranking model with non-linear kernel amounts to learning a linear hyperplane, hence efficient. The definition of CWS is deferred to Section~\ref{subsec:cws}. The derivation of our model and the learning algorithm are elaborated in Section~\ref{subsec:learning}.

The main difference of our model and other ranking methods is that, when training the model, we split the candidates of each image into two groups: one with top-$k$ rankings and the other consisting of the remaining candidates. We only compare the candidate from the first subset and the one from the second subset, instead of comparing all pairs of candidates. On account of such constraints, our model can focus on obtaining a reliable prediction for only the top-$k$ candidates rather than learning to rank all the candidates. Also note that our partial ranking model is different from top-$k$ ranking models in information retrieval, which aims to provide an accurate ranking for each top-$k$ retrieval~\cite{niu2012top}. In our case, it is not necessary to provide an accurate ranking within the top-$k$ candidates in that, when utilizing the $k$ proposals for further processing, like recognition, we typically do not care about the orders of proposals.

\subsection{Contribution}
We make two technical contributions in the work. First, by observing that the broadly studied \svmrank\ usually over constrains the object proposal problem, we suggest a partial ranking algorithm which produces an accurate estimation for the top-$k$ candidates. We further show that the partial ranking model is equivalent to a large-margin method where the margin separates top-$k$ and the remaining. Second, non-linear kernels are investigated in this work. To accelerate the training/testing phase as well as mitigate the memory overhead, we propose to adopt a consistent weighted sampling~(CWS) scheme so that learning a kernelized ranking model amounts to learning a linear model with the output of CWS.

\section{Related Work}
\label{rew}
Most of works in the literature aim at producing reliable candidates (i.e., high recall) without suffering efficiency. To this end, either effective feature descriptors and efficient grouping strategies are developed, or fast ranking algorithms are employed.

For example, \cite{cheng2014bing} developed a binarized normed gradient feature. \cite{alexe2010object} proposed an innovative super-pixel straddling and combined with multiple other descriptors in a Bayesian framework to qualify how likely a window covered an object. EdgeBoxes proposed an efficient objectness measurement that based on the edge feature \cite{zitnick2014edge}. In \cite{chen2015improving}, the multi-threshold super-pixels straddling features were proposed to refine the box locations. Recently, deep convolutional neural networks have been used to generate proposals. Specifically, in \cite{erhan2014scalable}, MultiBox combined the features of boxes, e.g., location as one layer in deep networks and outputted the confidence score for the box in the last hidden layer followed by a sigmoid. \cite{sermanet2013overfeat} proposed a network named ``OverFeat'' to extract powerful features for proposal generation.  \cite{ren2015faster} proposed region proposal networks which shared the convolutional feature maps used by detector for proposal boxes generation.

Another line considers different grouping strategies. For instance, Rigor proposed a piecewise-linear regression tree for grouping \cite{humayunrigor}. Selective Search (SS) merged adjacent super-pixels based on the similarity measurements including colour, texture and size \cite{uijlings2013selective}. Randomized Prim's (RP) employed similar features with Selective Search and merged super-pixels based on a randomized version of Prim's algorithm \cite{manen2013prime}. Multiscale Combinatorial Grouping (MCG) proposed a hierarchical segmenter that generated a tree of regions at multiple levels of homogeneity in brightness, color and texture. Then MCG trained a classifier to estimate the probability of object box boundary \cite{arbelaez2014multiscale}.

More in line with this work is the ranking algorithm. As the proposal generation methods usually generate thousands of proposals per image, ranking methods are utilized to select top-$k$ (in terms of the confidence score) candidates, where $k$ is usually a small quantity, e.g., $k = 500$. For comparison, we note that the potential boxes produced by sliding window can be up to millions. The ranking algorithms can be mapped to two major groups: those following the pairwise preferences~\cite{burges2005learning,joachims2002optimizing}, and the others are based on pointwise (e.g.,) classification~\cite{Proc:McRank_NIPS07}. The most popular scheme embedded with proposal generation methods is \svmrank. For example, Constrained Parametric Min-Cuts (CPMC) used the feature of RGB color distribution to form foreground super-pixel seeds. Then CPMC trained a regression model to rank all proposals based on a superset of 34-dimension features in total, including graph, region and Gestalt properties \cite{carreira2010constrained}. Geodesic Object Proposals (GOP) computed the static features such as location and adaptive features such as distance for segments and trained a linear ranking classifier to detect whether the boxes contained objects \cite{krahenbuhl2014geodesic}. \cite{zhang2011proposal} proposed to train a cascade of \svmrank $\ $ on gradient features. In the second stage, \svmrank \ learned the rank for all proposals. We also note that there is a large body of works for top-$k$ ranking algorithms in the community of information retrieval~\cite{xia2009statistical, ChenS15}. However, those approaches are not suitable for the objectiveness task, since in our problem, we do not have to assign a precise order for each of the top-$k$ proposals.

\section{Problem Formulation}
In this paper, we focus on a new re-ranking method for object proposal. Suppose that for each image, we have an ensemble of candidates (i.e., bounding boxes) ${B} = \{b_1, b_2, \cdots, b_n\}$\footnote{For simplicity, we assume that each image has a number of $n$ proposals.}. Each ensemble $B$ is associated with a vector $\by = \{y_1, y_2, \cdots, y_n\}$, with each $y_i$ being the IoU to the ground truth of the candidate $b_i$. Denote the input space as
\begin{equation}
\mathcal{X} = \{X \mid X = (\bx_1, \bx_2, \cdots, \bx_n) \},
\end{equation}
where $\bx_i \in \Rd$ is some feature descriptor for $b_i$ (e.g., HOG, CNN feature),
and the output space as
\begin{equation}
\mathcal{Y} = \{\by \mid \by = (y_1, y_2, \cdots, y_n) \}.
\end{equation}
We are interested in learning the prediction function
\begin{equation}
f:\quad \mathcal{X} \rightarrow \mathcal{Y}.
\end{equation}
To be more detailed, we assume that the function is parameterized by $\bw$
\begin{equation}
y_i = \bw \cdot \phi(\bx_i),\ \forall\ i =1, \cdots n,
\end{equation}
where $\bw$ is the weight vector we aim to learn, ``$\cdot$'' denotes the inner product and the potential $\phi(\bx)$ maps $\bx$ to a new feature space (which can be of infinite dimension). For linear separators, we know that $\phi(\bx)$ is an identical mapping. In this way, the mapping function $f$ is formulated as follows:
\begin{equation}
\label{eq:f(X)}
f(X; \bw) = (\bw \cdot \phi(\bx_1), \cdots, \bw \cdot \phi(\bx_n)).
\end{equation}

In the sequel, we will elaborate the design of the feature descriptor, the potential $\phi(\bx)$ and the learning algorithm for the weight vector $\bw$.

\subsection{Feature}
\label{subsec:feature}
The efficiency is one of most important attributes of a successful proposal algorithm. Thus we handcraft some simple features for computational efficiency. Given a bounding box $b$, its feature descriptor used here is the well known HOG feature~\cite{dalal2005histograms}. One may extract more effective features, such as those produced by convolutional neural networks (CNNs), which we have investigated through experiments in Section \ref{feature}. 

\subsection{Non-Linear Kernels and Consistent Weighted Sampling}
\label{subsec:cws}
In practice, instead of computing an explicit form of $\phi(\bx)$, one prefers to working with a reproducing Hilbert kernel space where for any vectors $\bu$ and $\bv$, there exists a kernel function $g(\bu, \bv)$ such that
\begin{equation}
g(\bu, \bv) = \phi(\bu) \cdot \phi(\bv).
\end{equation}
In this way, the kernel matrix $K = \{k_{ij} \mid k_{ij} = g(\bx_i, \bx_j)\}$ suffices for learning the parameter $\bw$. However, since the size of the kernel matrix is $n \times n$, the memory footprint always hinders an efficient learning.

In this paper, we alleviate the memory bottleneck of kernel matrix by utilizing a consistent weighted sampling scheme. In particular, we are interested in the so-called  min-max kernel \cite{cws,Proc:Ioffe_ICDM10,min-max,Report:Li_GMM16}. Formally, for any $d$-dimensional vectors $\bu$ and $\bv$ with non-negative components, it is defined as follows:
\begin{equation}
\label{eq:min-max}
\textrm{Min-Max:}\quad g_{mm}(\bu, \bv) = \frac{\sum_{i=1}^{d} \min\{u_i, v_i\}}{\sum_{i=1}^{d} \max\{u_i, v_i\}},
\end{equation}
where $u_i$ and $v_i$ denote the $i$th component of $\bu$ and $\bv$ respectively.

\vspace{0.1in}
\noindent {\bf Intuition.} To intuitively understand the min-max kernel, let us consider a special case where the element of $\bu$ and $\bv$ is either 0 or 1. In this way, one may verify that the numerator counts the number of non-zero entries of the set $\bu \cap \bv$ while the denominator explains that of the set $\bu \cup \bv$. Here, $\cap$ and $\cup$ denotes the element-wise ``and'' and ``or'' operation respectively. That is, for binary vectors, the min-max kernel computes their resemblance which is broadly used in information retrieval. Therefore,~\eqref{eq:min-max} essentially generalizes the well-known resemblance kernel for binary vectors to real-valued vectors with non-negative components. Since we are interested in the recall at top-$k$ which also can be viewed as information retrieval, we employ the min-max kernel in the work. For computational efficiency, we adopt the (0-bit) consistent weighted sampling technique~\cite{cws,Proc:Ioffe_ICDM10,min-max} which approximates the min-max kernel by linear functions, described in Algorithm~\ref{alg:cws}.
\begin{algorithm}[h]
\caption{Consistent Weighted Sampling~(CWS)}
\label{alg:cws}
\begin{algorithmic}[1]
\REQUIRE Feature vector $\bu \in \Rd$ with non-negative elements, number of trials $S$.
\ENSURE Consistent uniform samples $(i^*_1, i^*_2, \cdots, i^*_S)$ and $(t^*_1, t^*_2, \cdots, t^*_S)$.
\FORALL{$s = 1, 2, \cdots, S$}
\FORALL{$i = 1, 2, \cdots, d$}
\STATE $r_i \sim Gamma(2, 1),\quad c_i \sim Gamma(2, 1),\quad \beta_i \sim Uniform(0, 1)$.
\STATE $t_i \leftarrow \lfloor \log(u_i) / r_i + \beta_i \rfloor,\quad y_i \leftarrow \exp(r_i(t_i - \beta_i)),\quad a_i \leftarrow c_i / (y_i\exp(r_i))$.
\ENDFOR
\STATE $i^*_s = \argmin_i a_i,\quad t^*_s = t_{i^*_s}$.
\ENDFOR
\end{algorithmic}
\end{algorithm}

\begin{theorem}[\cite{Proc:Ioffe_ICDM10,min-max}, collision probability]\label{thm:1}
For any two non-negative vectors $\bu$ and $\bv$, let $(i^*_{s, \bu}, t^*_{s, \bu})$ and $(i^*_{s, \bv}, t^*_{s, \bv})$ be the consistent samples produced by Algorithm~\ref{alg:cws} at the $s$-th trial. Then we have
\begin{equation}
Pr\big( (i^*_{s, \bu}, t^*_{s, \bu}) = (i^*_{s, \bv}, t^*_{s, \bv}) \big) = g_{mm}(\bu, \bv).
\end{equation}
\end{theorem}

Due to the above theorem, we immediately have the following result:
\begin{equation}\label{eq:thm2}
\E\big[ 1\{(i^*_{s, \bu}, t^*_{s, \bu}) = (i^*_{s, \bv}, t^*_{s, \bv})\} \big] = g_{mm}(\bu, \bv),
\end{equation}
where the indicator function $1\{ \textrm{event} \}$ outputs 1 if event happens and 0 otherwise.

Eq.~\eqref{eq:thm2} implies that when we have sufficient number of consistent samples by Algorithm~\ref{alg:cws}, by comparing the identity for two vectors we obtain an estimation for the Min-max kernel.

\vspace{0.1in}
\noindent {\bf From Min-max kernel to linear mapping.} In order to show that \eqref{eq:thm2} virtually implies the min-max kernel~\eqref{eq:min-max} can be approximated by linear functions, let us consider an illustrative example. Suppose that the feature dimension $d = 256$ so that $i^*_s$ is bounded by 256. Suppose that $t^*_s$ is uniformly bounded by a constant $M$. We first map the pair $(i^*_s, t^*_s)$ to an integer $q = 256 * t^*_s + i^*_s$. Note that such mapping is injective since $i^*_s = (q \mod 256)$ and $t^*_s = \lfloor q / 256 \rfloor$. We further map the integer $q$ to a $(256M)$-dimensional binary vector with all but the $q$th entry equal to 1. We write the resultant sparse vector as $\bz$. In this way, we conclude that $1\{(i^*_{s, \bu}, t^*_{s, \bu}) = (i^*_{s, \bv}, t^*_{s, \bv})\} = \bz_{\bu} \cdot \bz_{\bv}$. Since we have $S$ number of consistent samples for both $\bu$ and $\bv$, we have
\begin{align*}
g_{mm}(\bu, \bv) = \E\big[ 1\{(i^*_{s, \bu}, t^*_{s, \bu}) = (i^*_{s, \bv}, t^*_{s, \bv})\} \big] \approx \frac{1}{S} \sum_{s=1}^{S} \bz_{s, \bu} \cdot \bz_{s, \bv} = \frac{1}{S} \bar{\bz}_{S, \bu} \cdot \bar{\bz}_{S, \bv},
\end{align*}
where $\bar{\bz}_{S, \bu} = ( \bz_{1, \bu}, \bz_{2, \bu}, \cdots, \bz_{S, \bu} )$ and likewise for $\bar{\bz}_{S, \bv}$. In other words, we can explicitly compute the feature mapping $\phi$ for the min-max kernel as above. More importantly, the mapped features are highly sparse which facilitates an efficient computation. An even simpler implementation is the ``0-bit'' strategy~\cite{min-max} by only using $q =i^*_s$ (i.e., using 0-bit from $t^*_s$). In our experiments, we find both implementations produce essentially indistinguishable results and hence we adopt the simpler version.

\subsection{Partial Ranking Model}
Given a training set $\{ (X^j, \by^j) \}_{j=1}^N$ where $N$ denotes the number of training images, we aim to learn the weight vector $\bw$, such that the top-$k$ candidates in each $X_j$ is better than the others. Assume without loss of generality that $\by_j = (y_j^1, \cdots, y_j^n)$ is in a non-ascending order. Let $\phi(\cdot)$ be the feature map for the min-max kernel as we discussed in the proceeding section. Thus, we are to solve the following convex optimization problem:
\begin{equation}
\label{eq:org svm}
\begin{split}
\min_{\bw}\quad& \frac{1}{2} \Vert \bw \Vert_2^2,\\
\st\quad & \bw \cdot \phi(\bx^j_p) \geq \bw \cdot \phi(\bx^j_q),\ \forall\ j \in [N], p \in [k], q \in [n] \backslash [k],
\end{split}
\end{equation}
where $[N]$ denotes the integer set of $\{1, \cdots, N\}$ and likewise for $[n]$ and $[k]$. Note that we presume the bounding boxes (and hence $\bx$'s) are arranged in a non-ascending order. Hence, the above program splits for each image the bounding boxes into two subsets: one with the $k$ largest IoU with ground truth and the other is the remaining. The partial ranking model comes from such kind of constraints since we only impose that candidates from the first subset is better than the one from the second one.

In previous works, a commonly utilized ranking model is \svmrank, which is formulated as follows:
\begin{equation}
\label{eq:rank svm}
\begin{split}
\min_{\bw}\quad& \frac{1}{2} \Vert \bw \Vert_2^2,\\
\st\quad & \bw \cdot \phi(\bx^j_p) \geq \bw \cdot \phi(\bx^j_q),\ \forall\ j\in[N],\ p, q \in[n],\ p \leq q.
\end{split}
\end{equation}
The prominent difference is that the set of constraints in our formulation~\eqref{eq:org svm} is a subset of \svmrank. Our formulation is motivated by the practical usage of object proposals: in principle, proposals alleviate the shortcoming of the sliding window scheme by reducing the search space from the whole image (over each position and scale) to a manageable number of regions (i.e., bounding boxes). Based on these candidates, one may extract finer visual cues for accurate recognition and detection. Thus, essentially a good prediction for the top-$k$ candidates is sufficient for a successful proposal algorithm. From this point of view, it is more reasonable to compare the top-$k$ candidates (which are of interest for the user) to the remaining $n-k$ ones, rather than comparing any pair from the ensemble of the candidates (which is the formulation of~\eqref{eq:rank svm}). For practitioners, our formulation is more appealing since it reduces the number of constraints of \svmrank\ from $O(n^2)$ to $O(nk)$, and hence a more efficient learning procedure.

\subsection{Learning with Large Margin Model}
\label{subsec:learning}
In practical problems, the ideal constraints in \eqref{eq:org svm} might be violated owing to dirty data, improper features etc. Thus it is necessary to derive a soft-margin formulation. We discuss two variants to~\eqref{eq:org svm}.

\vspace{0.1in}
\noindent {\bf Variant 1.} The first proposed variant resembles binary SVM. First, we transform \eqref{eq:org svm} to the hard large margin based model:
\begin{equation}
\label{eq:hard svm}
\begin{split}
\min_{\bw}\quad& \frac{1}{2} \Vert \bw \Vert_2^2,\\
\st\quad & \bw \cdot \phi(\bx_j^p) \geq +1,\forall \ p\in[k], \forall\ j \in [N],\\
&\bw \cdot \phi(\bx_j^q) \leq -1,\forall \ q\in[n]\backslash[k], \forall\ j \in [N].
\end{split}
\end{equation}
The above formulation is ``equivalent'' to Problem~\eqref{eq:org svm} in the sense that the partial ordering of top-$k$ and the remaining is preserved. In other words, the reformulation keeps the relative ranking for the subsets $[k]$ and $[n]\backslash [k]$.

By introducing non-negative slack variables $\xi_j$, we obtain the soft margin formulation:
\begin{equation}
\label{eq:soft svm}
\begin{split}
\min_{\bw, \xi_1, \dots, \xi_N}\quad& \frac{1}{2} \Vert \bw \Vert_2^2 + C \sum_{j=1}^{N} \xi_j,\\
\st\quad & \bw \cdot \phi(\bx_j^p) \geq +1 - \xi_j,\ \forall \ p\in [k], \forall\ j \in [N],\\
&\bw \cdot \phi(\bx_j^q) \leq -1 + \xi_j,\ \forall \ q\in [n]\backslash[k], \forall\ j \in [N],
\end{split}
\end{equation}
where $C$ is a non-negative trade-off parameter.

We remark here the main difference with the binary SVM. For binary SVM, the positive and negative samples are usually picked by some binary evaluation metric. For example, for image classification, a sample is considered to be positive if it contains some object. There is no \emph{ranking} information in binary SVM. In our case, we actually rank the samples and select the top-$k$ candidates as positive.

\vspace{0.1in}
\noindent {\bf Variant 2.} The second variant is more involved as follows:
\begin{equation}
\label{eq:var2}
\begin{split}
\min_{\bw, \bx i_1, \dots, \bx i_N}\quad& \frac{1}{2} \Vert \bw \Vert_2^2 + C \sum_{j=1}^{N} \xi_j,\\
\st\quad & \bw \cdot \big[  \phi(\bx^j_p) - \phi(\bx^j_q) \big]\geq 1 - \xi_j,\ \forall\ j\in[N], \forall\ p \in [k], \forall\ q \in [n]\backslash[k].
\end{split}
\end{equation}
The difference to Variant 1 is that the above program explicitly encodes ranking information while that of Variant 1 is implicitly imposed. Note that by defining new training samples as $\{ \phi(\bx^j_p) - \phi(\bx^j_q) \}$ over $j \in [N]$, $p \in [k]$ and $q \in [n]\backslash [k]$, and assigning positive labels to them, we can utilize linear SVM (e.g., LibLinear) to learn the above model. In our experiments, we adopt Variant 2 for the proposal ranking.

\newpage\clearpage

\section{Experiments}
\label{sec:exp}

\noindent {\bf Dataset.}
We evaluate our method on the popular PASCAL Visual Object Challenge (VOC) 2007 dataset~\cite{everingham2010pascal}. The dataset contains 9,963 images belonging to 20 classes and has standard subset for training, validation and testing. We use the first two subsets for our training.  

\vspace{0.1in}
\noindent{\bf Evaluation Metrics.} \
The employed evaluation metrics are Recall, Average Recall (AR)~\cite{Hosang2015pami} and Mean Average Best Overlaps (MABO)~\cite{uijlings2013selective}. We say that a proposal contains an object if it is above a given intersection over union (IoU) threshold. Average recall is a recently proposed metric to evaluate proposals in a range of IoU threshold. Average recall also evaluates the contribution of proposals to detection performance. We report average recall for the threshold of IoU of the interval $[0.5, 1.0]$. MABO is defined as the mean of the Average Best Overlap over all classes while ABO is the best overlap between each ground truth box and generated proposals.

\vspace{0.1in}
\noindent{\bf Baselines.} \
We collect recently established proposal algorithms for comparison. To be more detailed, we compare with BING~\cite{cheng2014bing}, CPMC \cite{carreira2010constrained}, GOP \cite{krahenbuhl2014geodesic}, EB~\cite{zitnick2014edge}, Endres \cite{endres2010category}, MCG~\cite{arbelaez2014multiscale}, OBJ~\cite{alexe2010object}, Rigor \cite{humayunrigor}, Rantalankila \cite{mvg2014}, RS, M-MCG~\cite{chen2015improving}, RP~\cite{manen2013prime} and SS~\cite{uijlings2013selective}. Here, RS refers to regular sampling which is a fast scheme to produce a large number of candidates~\cite{cheng2014bing,zitnick2014edge}. To test the potential detection result based on the proposals, we also choose the state-of-the-art deep network based region box generation methods as baselines \cite{ren2015faster}, denoted as VGG and ZF. The recall and average recall results of all baseline methods with 300 proposals per image are listed on Table~\ref{tab:prop_300} and Table~\ref{tb:ar300}. The number of 300 proposals per image is suggested by \cite{ren2015faster}. The proposals generated by VGG and ZF on the VOC07 test dataset are downloaded from their public website \footnote{\url{https://github.com/ShaoqingRen/faster_rcnn}}.

\vspace{0.1in}
\noindent{\bf Our Algorithm.} \
Since the focus of the paper is on the ranking algorithm, any ensemble of proposals can be fed to our algorithm. Here, we illustrate the efficacy of partial ranking~(PR) with five choices: proposals produced by EB, GOP, MCG, OBJ, Rigor, RS and SS. Correspondingly, we name our algorithms as PR-EB, PR-MCG, PR-OBJ, PR-SS and PR-RS.

\subsection{Examine the Influence of Features and Min-Max Kernel}
\label{feature}
\begin{figure}[t]
	\subfloat[Recall at 0.5 IoU.]{
		\includegraphics[width=0.33\textwidth]{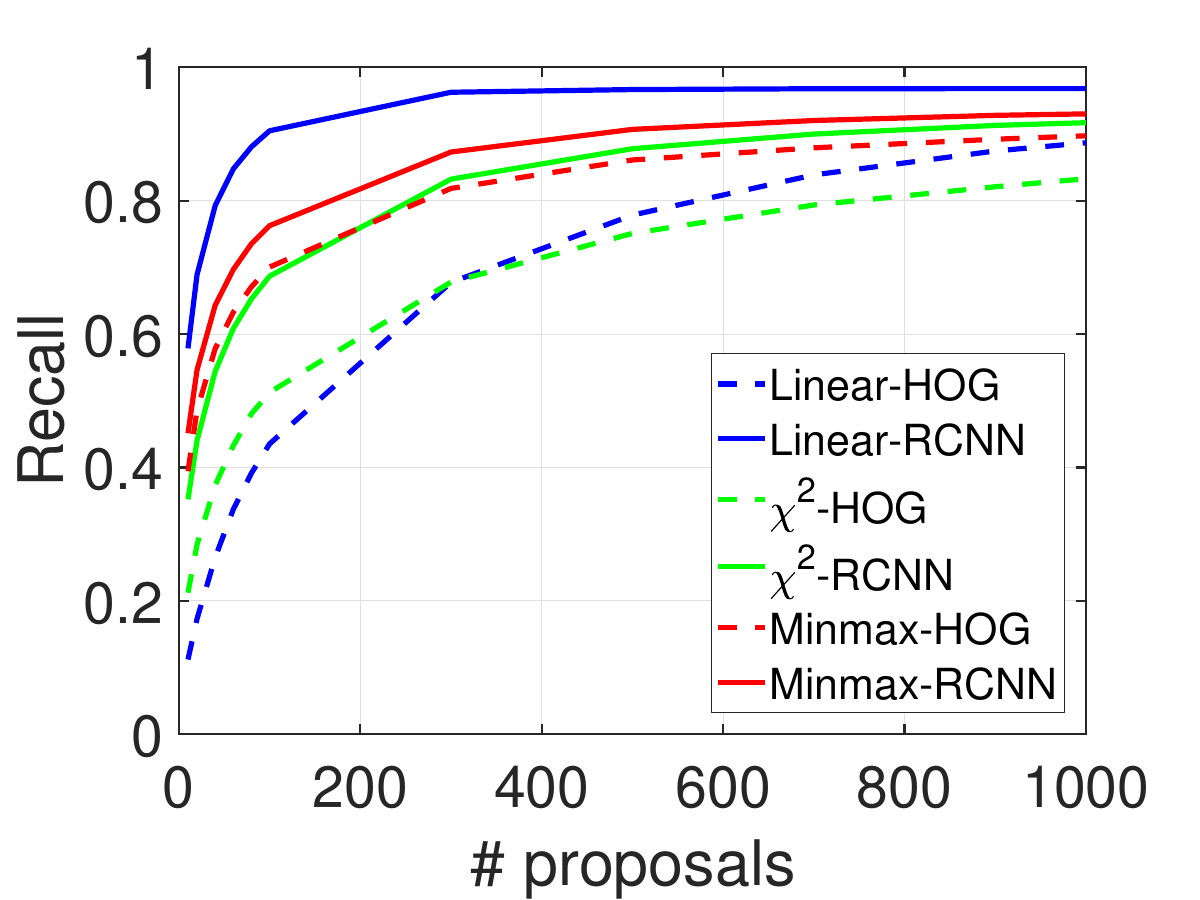}}
  \subfloat[Recall at 0.8 IoU.]{
  \includegraphics[width=0.33\textwidth]{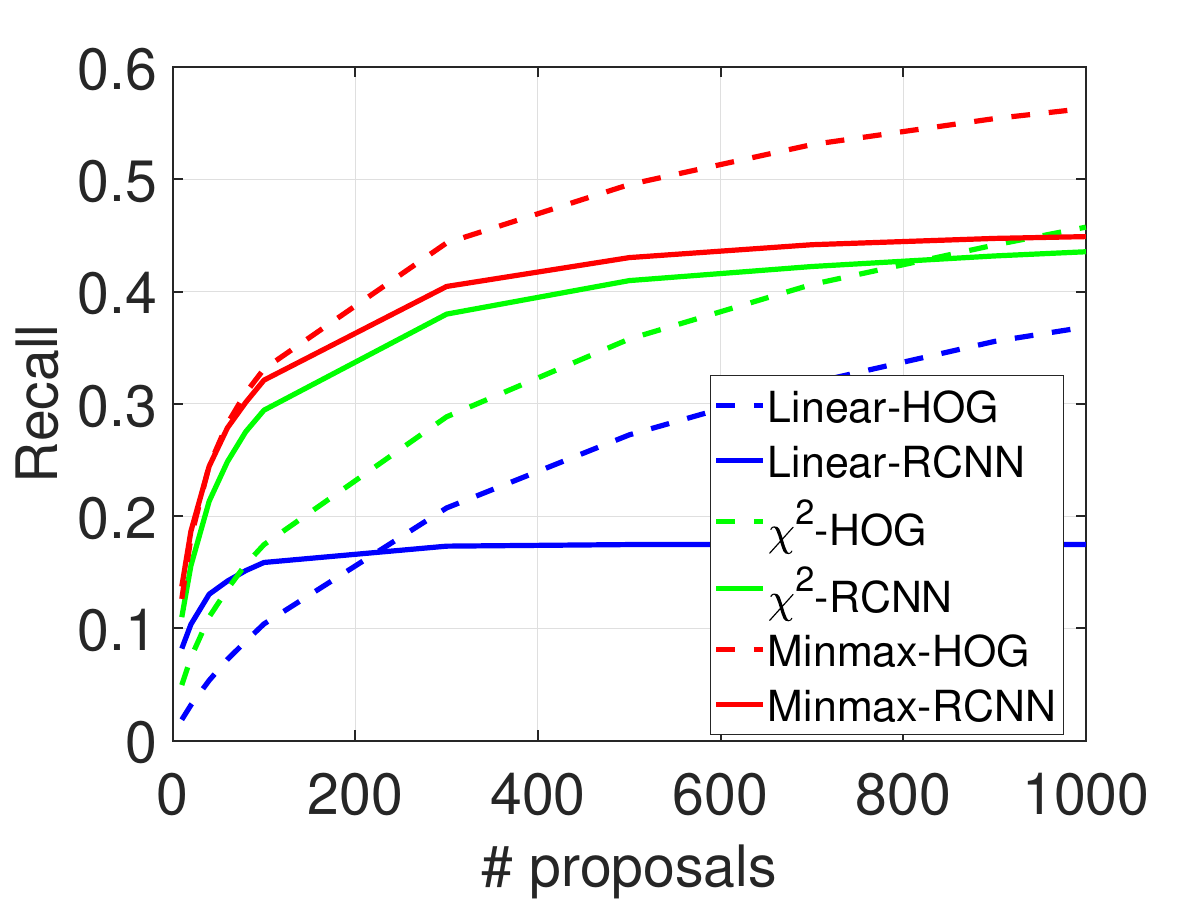}}
       \subfloat[Average recall.]{
     \includegraphics[width=0.33\textwidth]{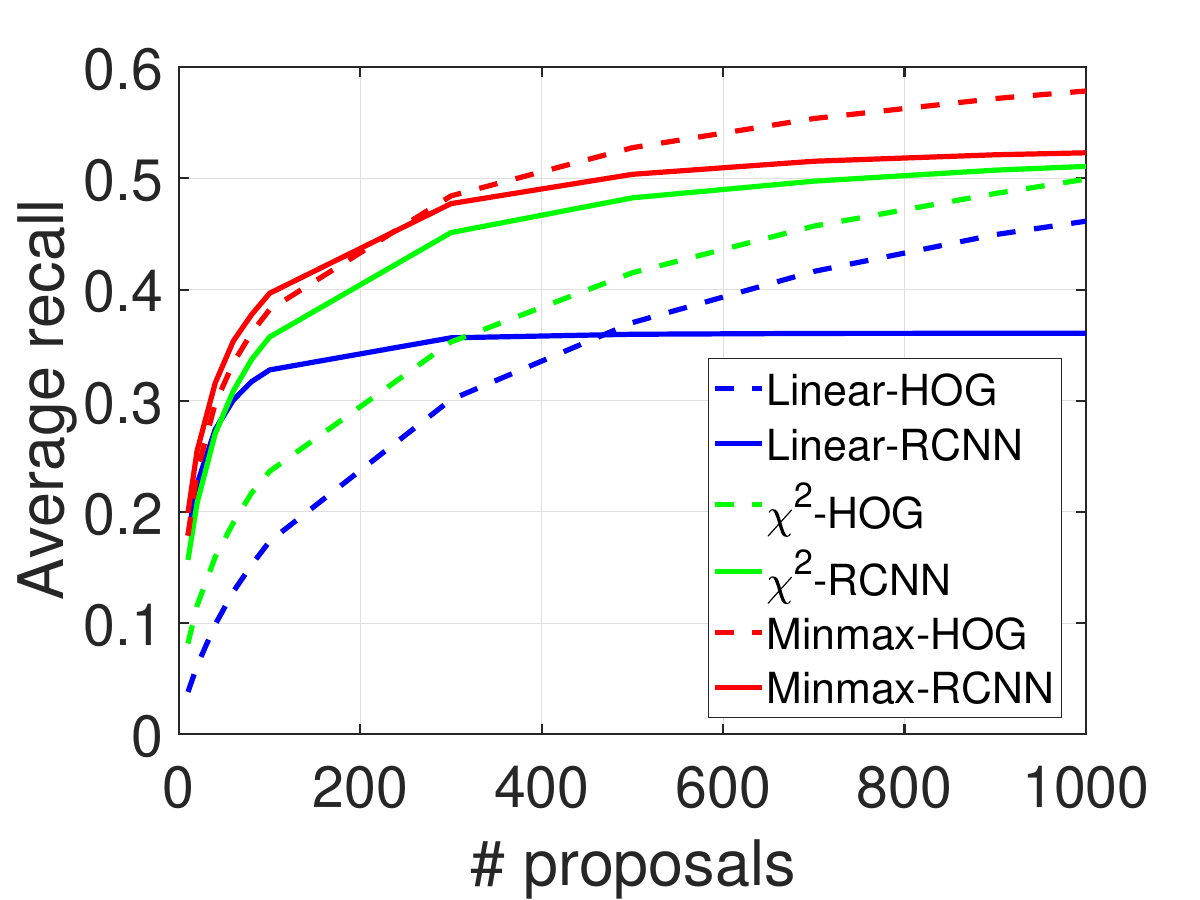}}
\caption{\scriptsize{Recall versus number of proposals compared with HOG and CNN feature, Linear and Min-Max Kernel}}
\label{fig:hog_rcnn_ar}
\end{figure}

Note that any proposal feature can be adopted to train our partial ranking model. In this section, we investigate the influence of various features towards the recall and average recall. The precomputed proposals are generated by Selective Search as an example. Here we choose two features: HOG and CNN-feature. For HOG, each proposal box is resized then  the feature is extracted by the public toolkit VLFeat with cell size of 8. The feature dimension of each proposal is 1488. For CNN feature, the fine tuned model trained on VOC07 train and validation set is utilized to extract the feature. The dimension of each proposal is 4096. Then we perform CWS on features with 1024 trials, i.e., $S = 1024$ for Algorithm~\ref{alg:cws}. The ranking model is trained with top $k=80$ proposals as positive. The partial ranking model with HOG and CNN features are denoted as Minmax-HOG and Minmax-RCNN. In the same way, the linear model with HOG and CNN features are denoted as Linear-HOG and Linear-CNN. We also compare our kernel with $\chi^2$ kernel~\cite{vedaldi2012efficient,Proc:Li_NIPS13}. The $\chi^2$ kernel equipped with HOG and CNN features are denoted as $\chi^2$-HOG and $\chi^2$-CNN. For a fixed number of proposals, we plot the recall for different IoU in Figure~\ref{fig:hog_rcnn_iou}. Figure~\ref{fig:hog_rcnn_ar}a and Figure~\ref{fig:hog_rcnn_ar}b show the recall when the number of proposals per image is varied in the testing phase. The average recall between IoU 0.5 to 1.0 versus the number of proposals is plot in Figure~\ref{fig:hog_rcnn_ar}c. From the figures, we can get the following observations.

\begin{itemize}
\item HOG feature vs. CNN feature. It is shown that Linear-CNN (solid blue curve) outperforms Linear-HOG (dashed blue curve) in recall and average recall in most cases. But Minmax-HOG (dashed red curve) achieves better recall than Minmax-CNN (solid red curve) when the IoU threshold is larger than 0.6 as shown in Figure~\ref{fig:hog_rcnn_iou}. Specifically, in the case of 300 proposals per image, the recall of Minmax-HOG is 49.66\%, while Minmax-CNN is 41.57\% with IoU 0.8.
\begin{figure}[h!]
	\subfloat[100 proposals per image.]{
		\includegraphics[width=0.33\textwidth]{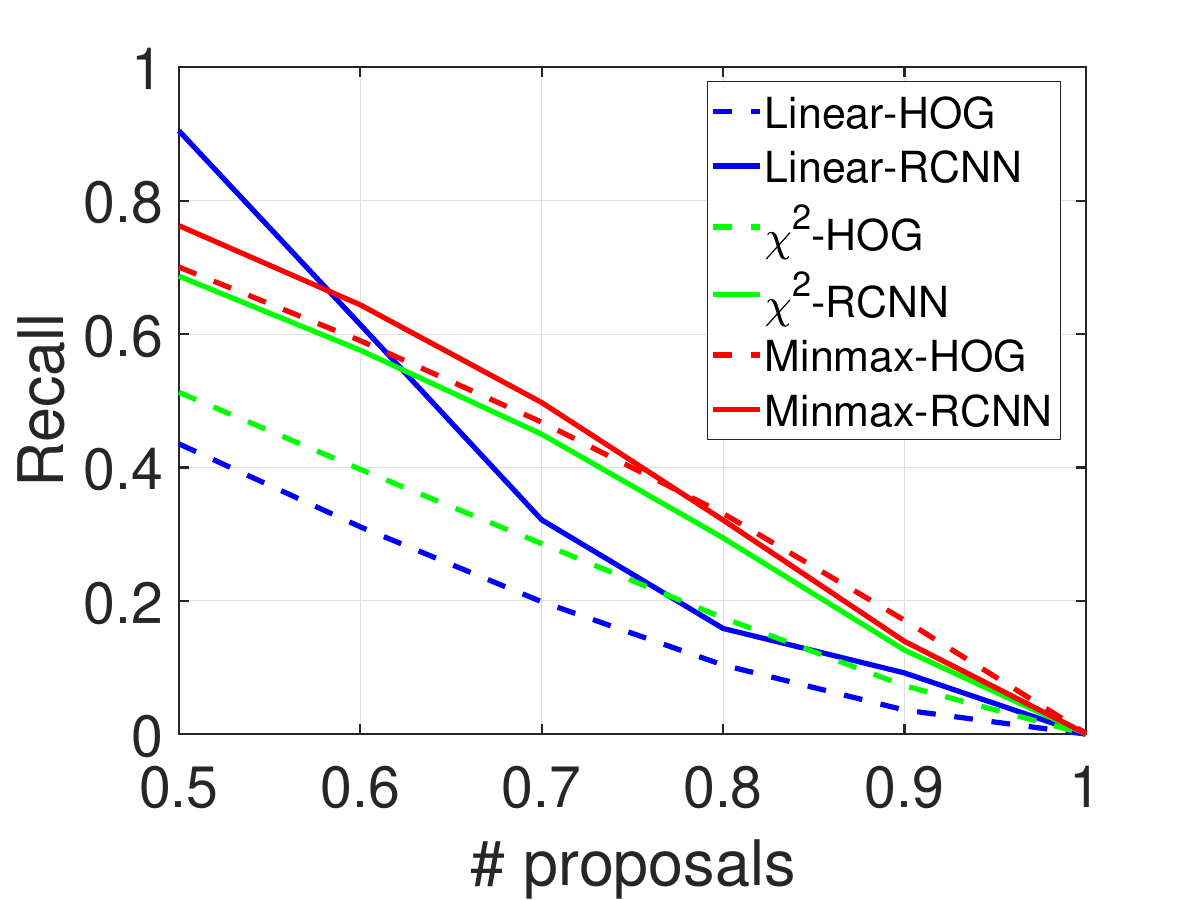}}
  \subfloat[300 proposals per image.]{
  \includegraphics[width=0.33\textwidth]{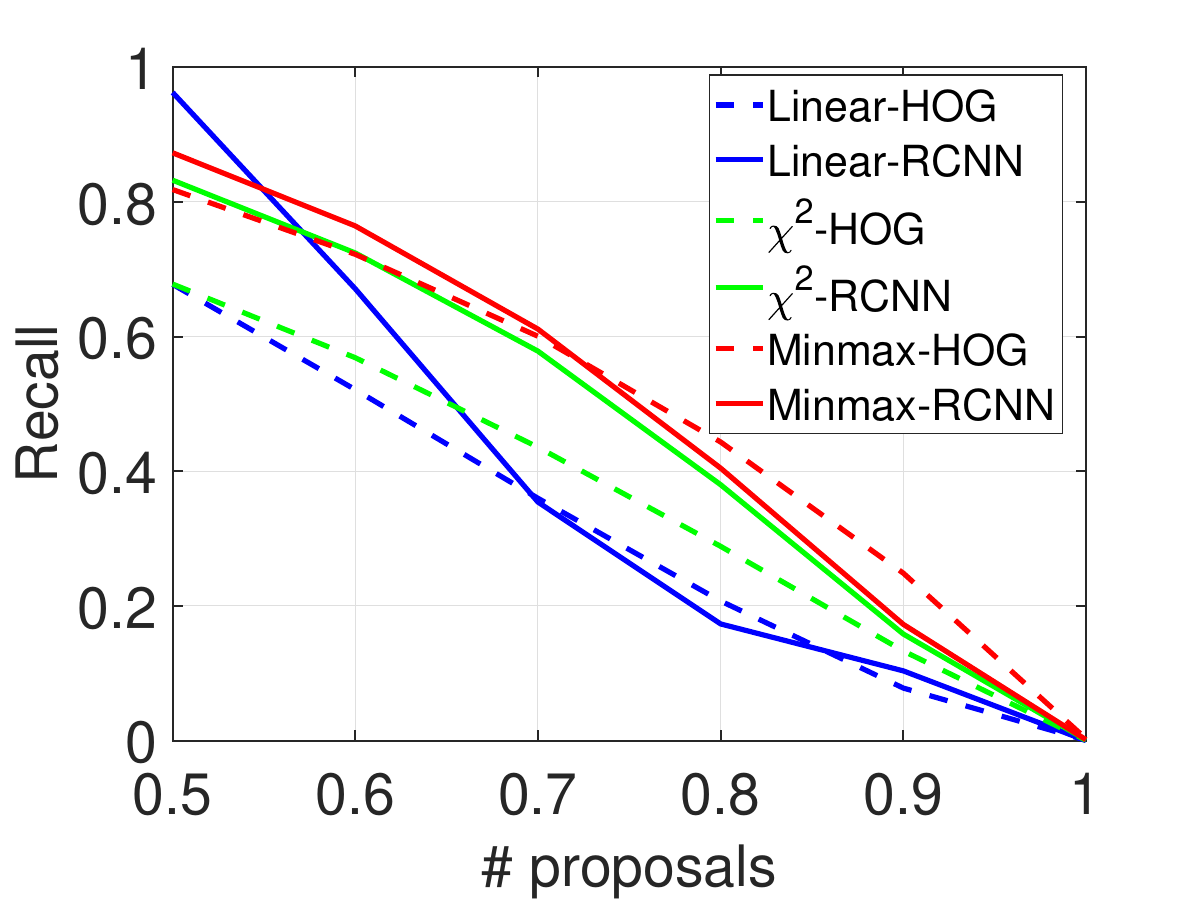}}
       \subfloat[500 proposals per image.]{
     \includegraphics[width=0.33\textwidth]{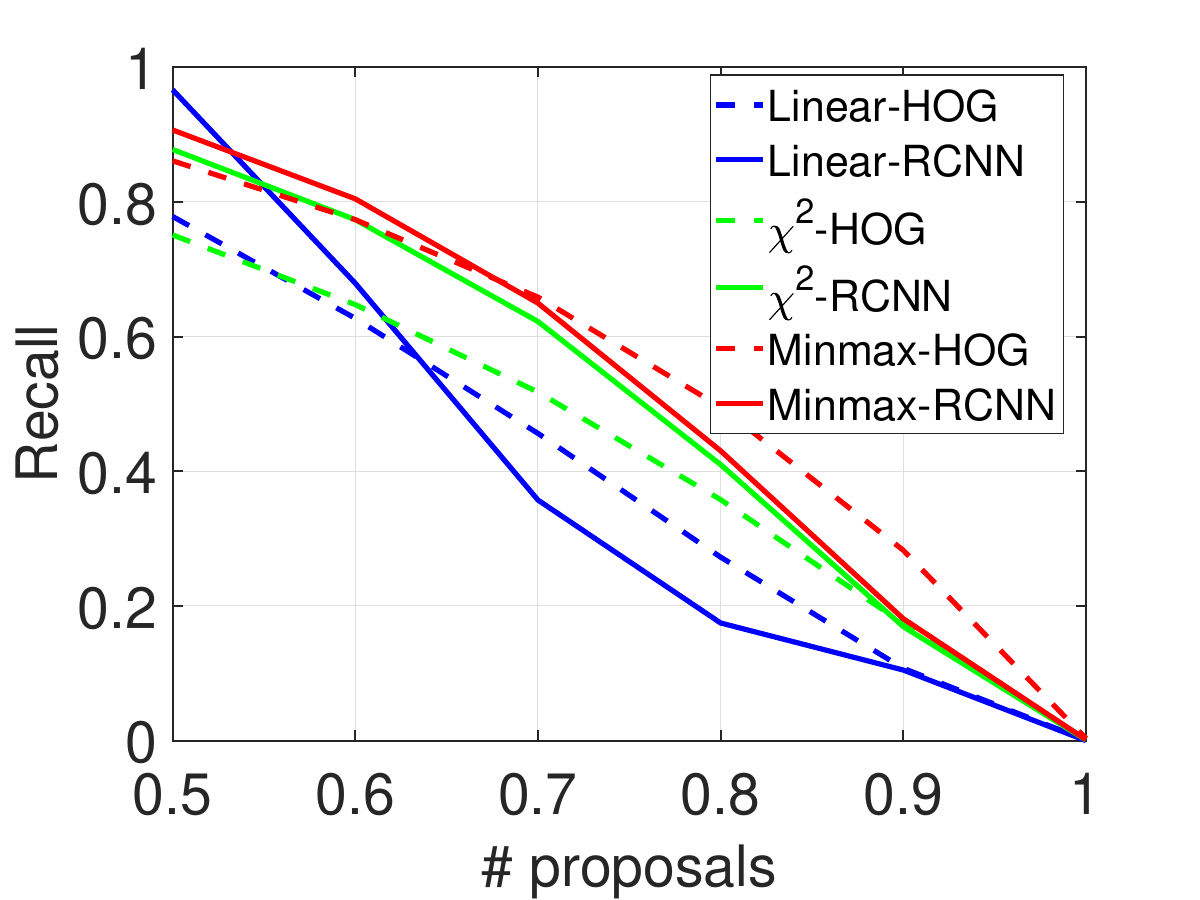}}
\caption{{Recall versus IoU threshold compared with HOG and CNN feature, Linear and Min-Max Kernel}}
\label{fig:hog_rcnn_iou}
\end{figure}
\item Linear kernel vs. Min-Max kernel. We can observe that the Min-Max models (red curves) always outperform the linear models (blue curves) no matter equipped with HOG or CNN features. For instance, in case of 1000 proposals per image with IoU threshold 0.8, the recall of Minmax-HOG is 60.10\%, while Linear-HOG is less than 40\%.
\item $\chi^2$ kernel vs. Linear kernel vs. Min-Max kernel. The $\chi^2$ kernel obtains better results in terms of recall and average recall than linear kernel. However, our Min-Max kernel still achieves great advantage over $\chi^2$ kernel based method.
\end{itemize}

In summary, different descriptors can be utilized for specific tasks. In our paper, the proposals are mainly generated for detection, which is highly correlated with average recall. As shown in Figure~\ref{fig:hog_rcnn_ar}c, Minmax-HOG achieves the best average recall. Moreover, HOG is more efficient and obtains better recall than CNN with tight IoU threshold. Thus, we choose HOG feature as default setting.

\subsection{Examine the Influence of $k$}
\begin{figure}[h!]
        \subfloat[\scriptsize{100 proposals per image, IoU=0.5.}]{
 \includegraphics[width=0.33\textwidth]{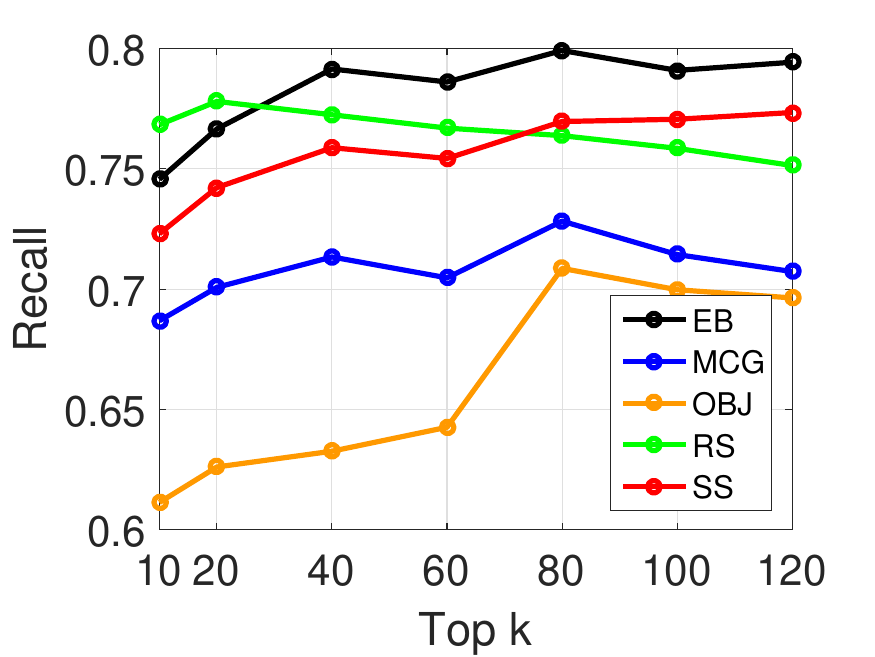}}
         \subfloat[\scriptsize{300 proposals per image, IoU= 0.6.}]{
  \includegraphics[width=0.33\textwidth]{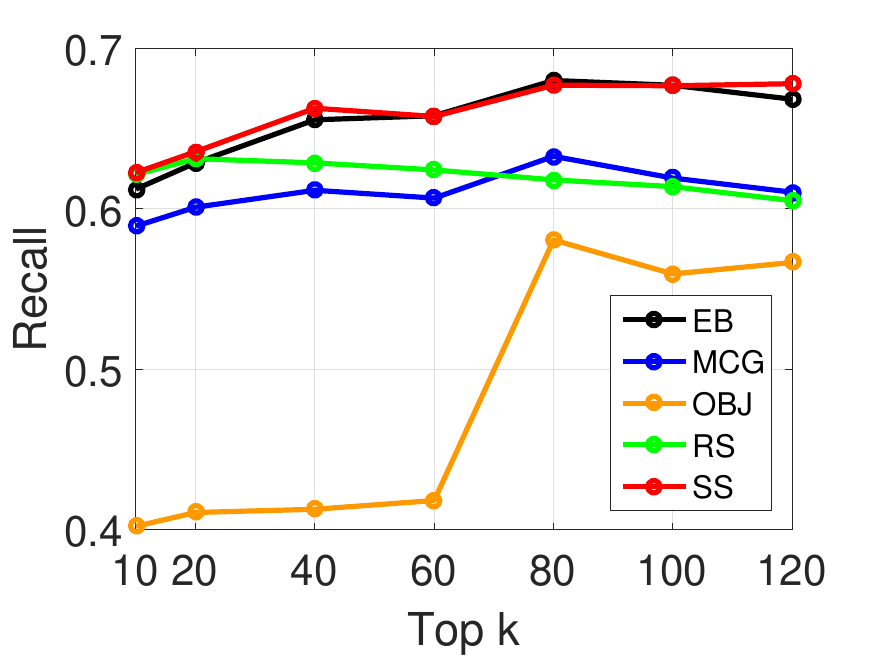}}
       \subfloat[\scriptsize{500 proposals per image.}]{
  \includegraphics[width=0.33\textwidth]{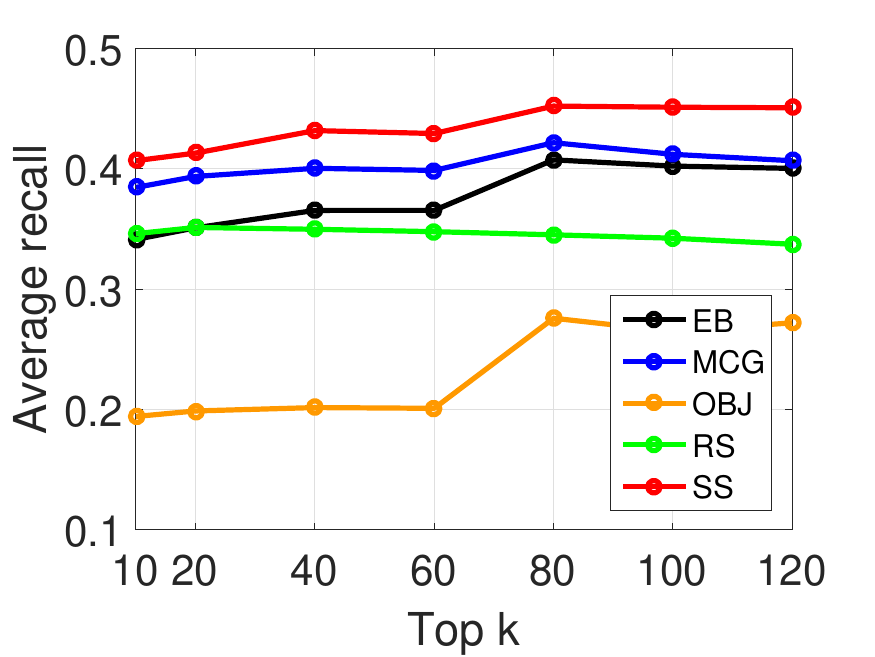}}
  \caption{{Recall versus $k$ compared with improved methods.}}
\label{fig:kall}
\end{figure}
\begin{figure}[t]
        \subfloat[100 proposals per image.]{
 \includegraphics[width=0.33\textwidth]{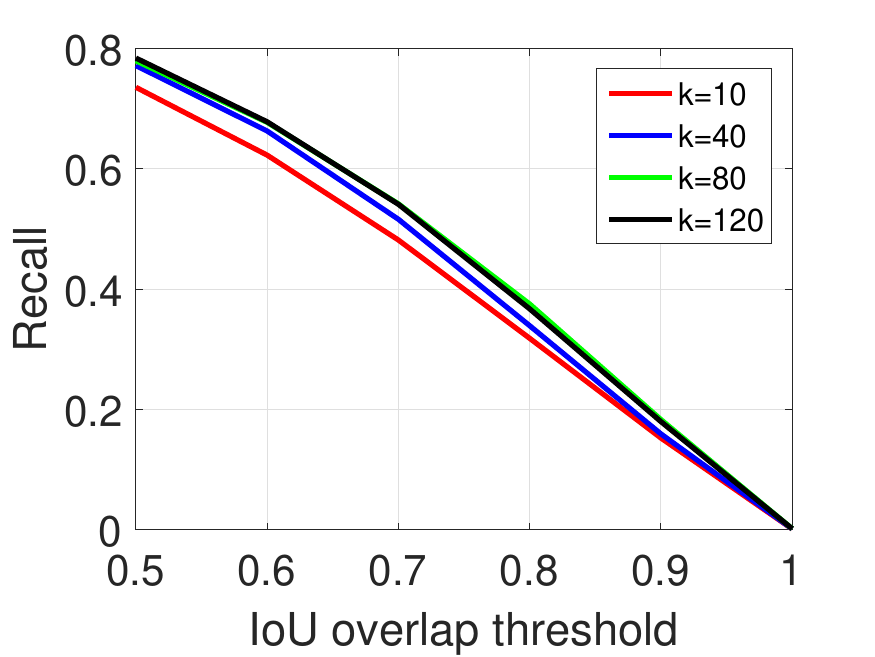}}
         \subfloat[Recall at 0.5 IoU.]{
  \includegraphics[width=0.33\textwidth]{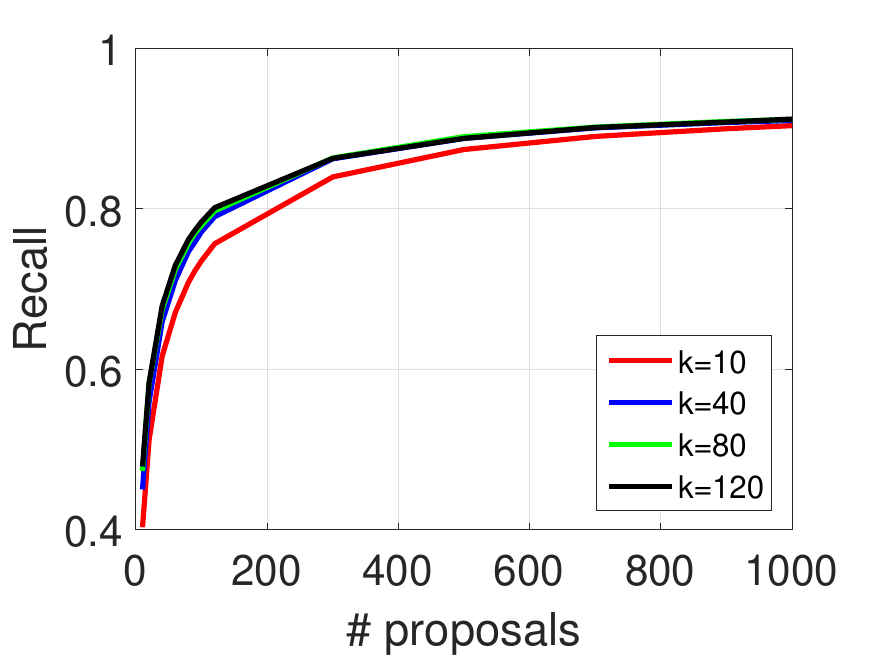}}
       \subfloat[Average recall.]{
  \includegraphics[width=0.33\textwidth]{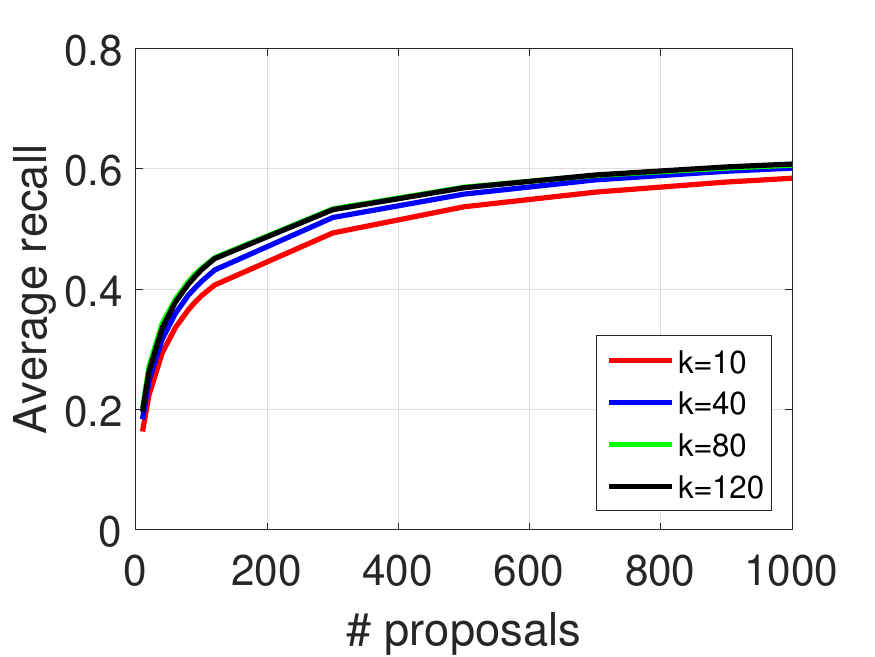}}
  \caption{{Recall versus IoU threshold, number of proposals compared with $k=[10,40,80,120]$.}}
\label{fig:kss}
\end{figure}

\begin{figure}[t]
	\subfloat[Recall at 0.5 IoU.]{
		\includegraphics[width=0.33\textwidth]{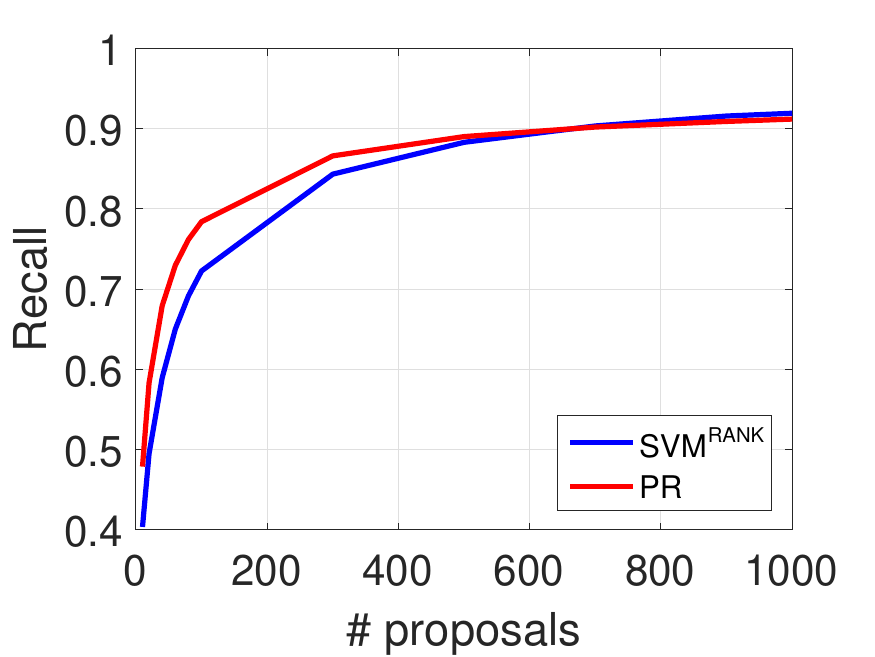}}
  \subfloat[Recall at 0.8 IoU.]{
  \includegraphics[width=0.33\textwidth]{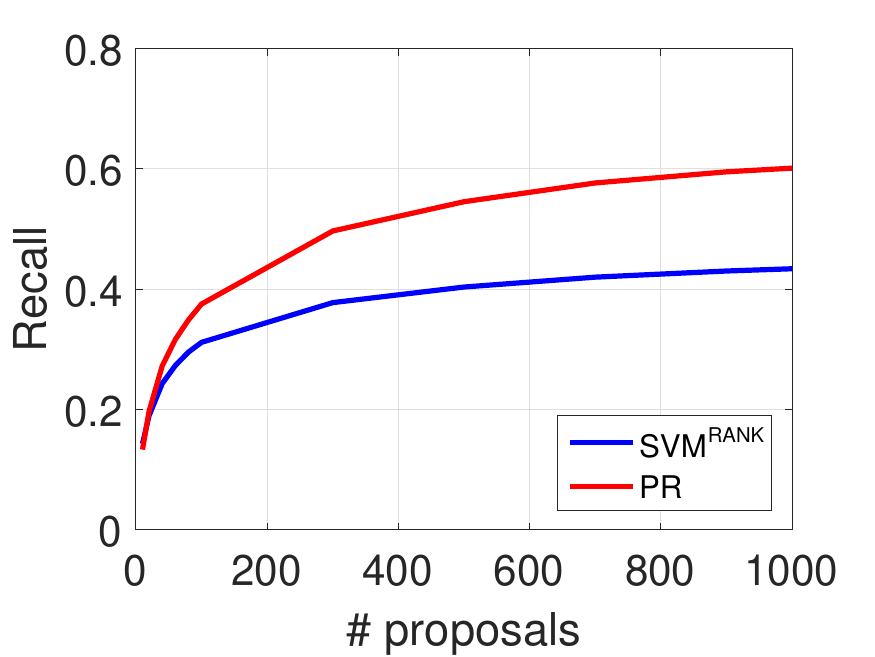}}
       \subfloat[Average recall.]{
     \includegraphics[width=0.33\textwidth]{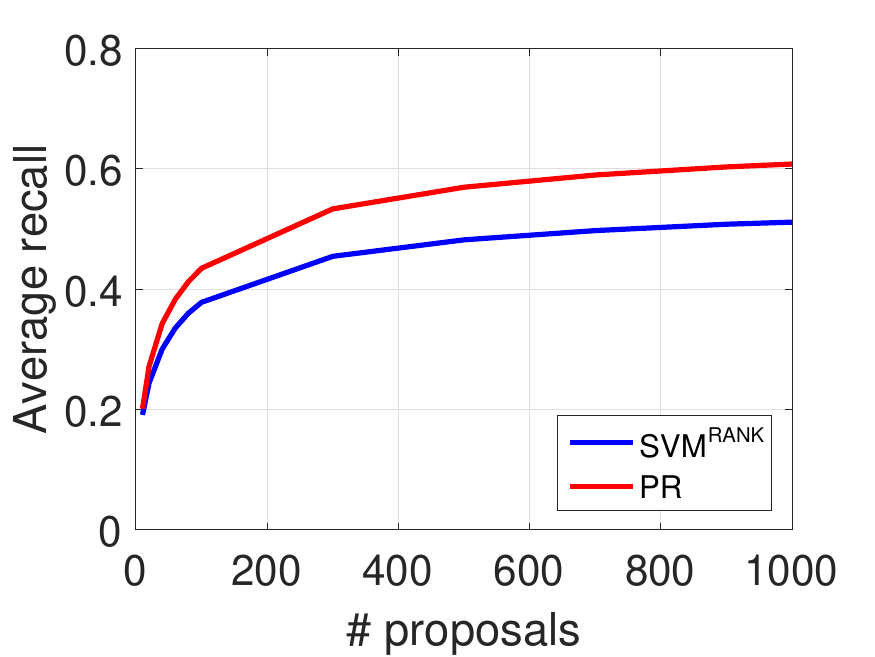}}
\caption{\textbf{Recall versus number of proposals compared with \svmrank~and partial ranking model.}}
\label{fig:pr_linear_ar}
\end{figure}
In this section, we investigate the influence of $k$ towards the recall and average recall. We set $n=2000$ in our experiments, i.e., there are 2000 candidates obtained from proposal generation algorithms for re-ranking. We tune the quantity of $k$ in the range of  $[10,20,40,60,80,100,120]$ in the training phase and examine how the recall changes with it. The proposals are provided by EB, MCG, OBJ, RS and SS. Figure~\ref{fig:kall} shows the recall and average recall with the increase of $k$. 
\begin{figure}[h!]
	\subfloat[100 proposals per image.]{
		\includegraphics[width=0.33\textwidth]{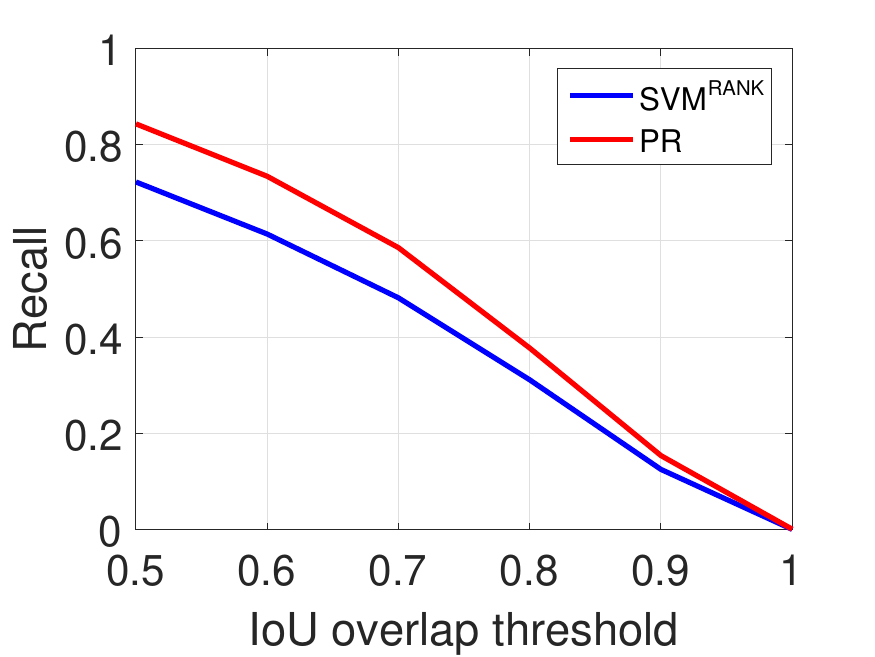}}
  \subfloat[300 proposals per image.]{
  \includegraphics[width=0.33\textwidth]{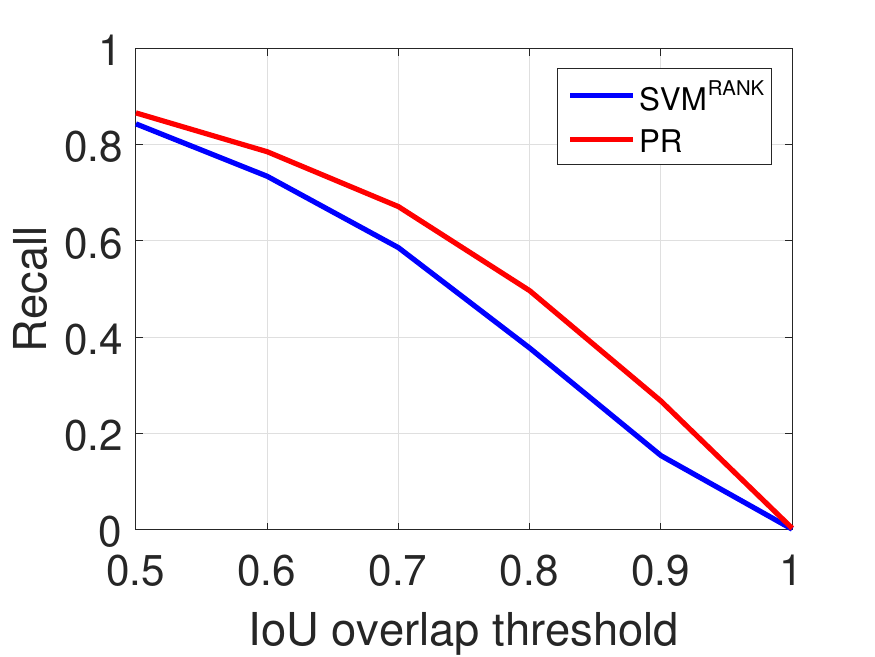}}
       \subfloat[500 proposals per image.]{
     \includegraphics[width=0.33\textwidth]{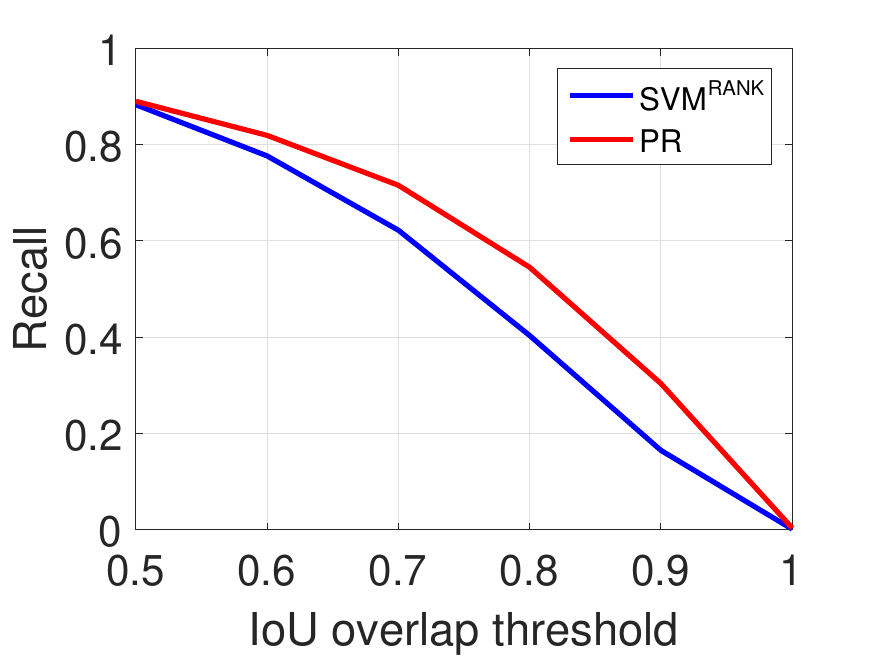}}
\caption{\textbf{Recall versus IoU threshold compared with \svmrank~and partial ranking model.}}
\label{fig:pr_linear_iou}
\end{figure}
 Interestingly, we find that our model generalizes well. Specifically, recall of all methods improves with the increase of $k$ but has little vibration when $k$ reaches 80. For instance, in case of 300 proposals per image, the recall of OBJ is 41.82\% ($k=60$), 58.06\% ($k=80$), and 55.94\% ($k=100$). But the improvement is not significant for some methods, such as Selective Search. Specifically, we choose the proposals from Selective Search and plot the recall versus $k\in [10,40,80,120]$ in Figure \ref{fig:kss}. We can observe that the recall still improves a little when $k$ increases from 40 to 120. For example, in case of 100 proposals per image with IoU threshold 0.5, the recall of SS reaches 77.15\%, 77.94\% and 78.40\% when $k$ is 40, 80 and 120 respectively.  Thus, the setting of $k=80$ is sufficient to train our model which also enjoys efficiency in memory and computation. In the following sections, the default setting of $k$ is 80.

\subsection{Examine the Effect of Partial Ranking}

Here we examine the effect of partial ranking~(PR) by comparing it with \svmrank. The proposals are provided by Selective Search as an example. We train the model by \svmrank\ and partial ranking model and present the recall in Figure \ref{fig:pr_linear_ar} and Figure \ref{fig:pr_linear_iou}. It is shown that partial ranking model is always superior to \svmrank~in terms of both recall and average recall. In case of IoU threshold 0.5 in Figure \ref{fig:pr_linear_ar}a, the two models achieve similar recall when the number of proposals is larger than 400. But with tight IoU threshold 0.8 in Figure \ref{fig:pr_linear_iou}, our partial ranking model favors a great advantage over \svmrank. Specifically, in case of 700 proposals per image, the recall of PR is 57.64\%, while \svmrank\ only reaches 41.98\%. Regarding average recall, our partial ranking model also has a dramatic advantage over \svmrank. For example, with 100 proposals per image, the partial ranking model obtains the average recall of 43.46\%, while \svmrank\ is 37.80\%. It demonstrates that the partial ranking model can locate proposals with higher accuracy compared with \svmrank\ model.

\subsection{Comparison with Baselines}

\begin{figure}[h!]
	\subfloat[100 proposals per image.]{
		\includegraphics[width=0.5\textwidth]{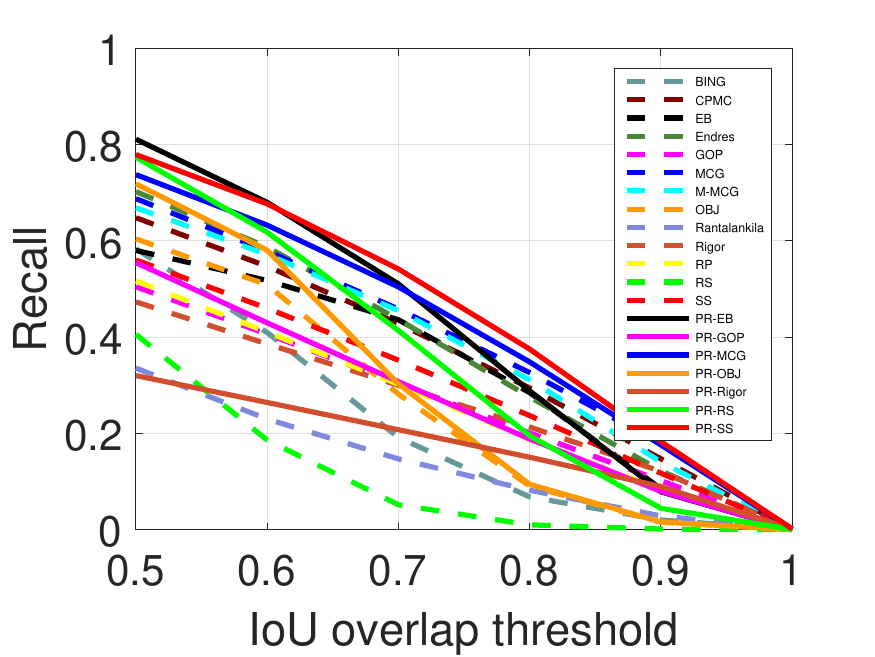}}
  \subfloat[500 proposals per image.]{
  \includegraphics[width=0.5\textwidth]{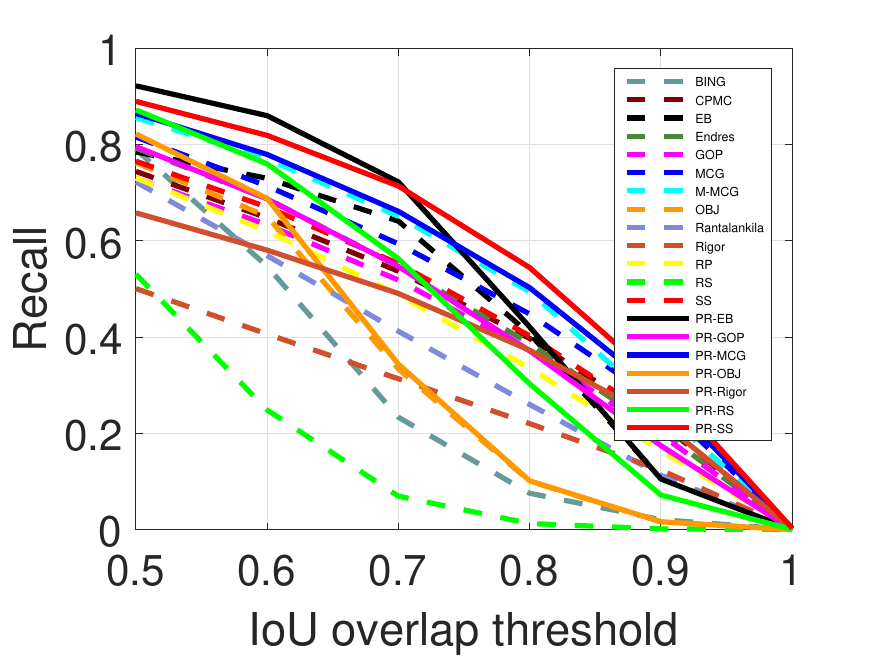}}

       \subfloat[1000 proposals per image.]{
    \includegraphics[width=0.5\textwidth]{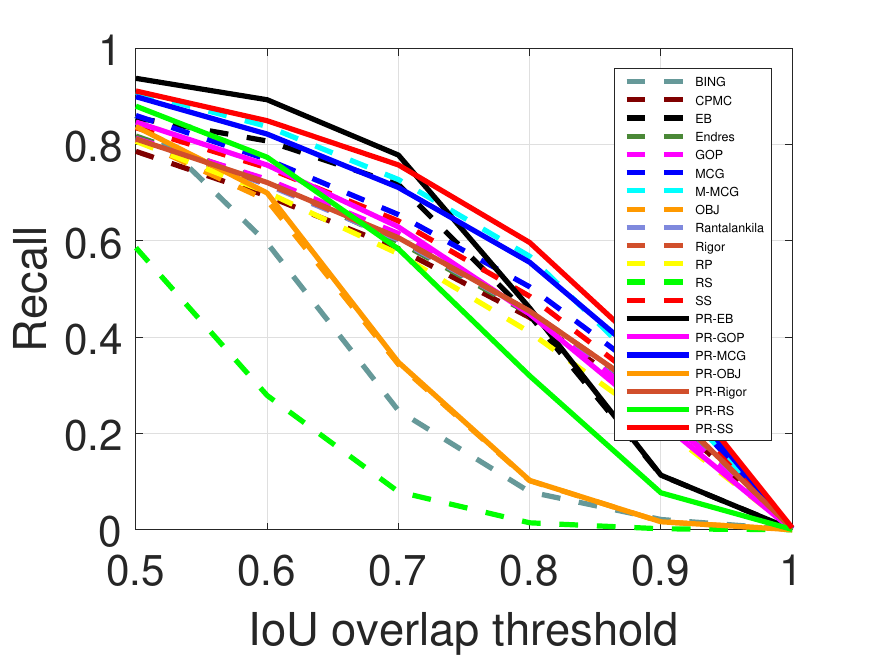}}
        \subfloat[Recall a 0.5 IoU.]{
 \includegraphics[width=0.5\textwidth]{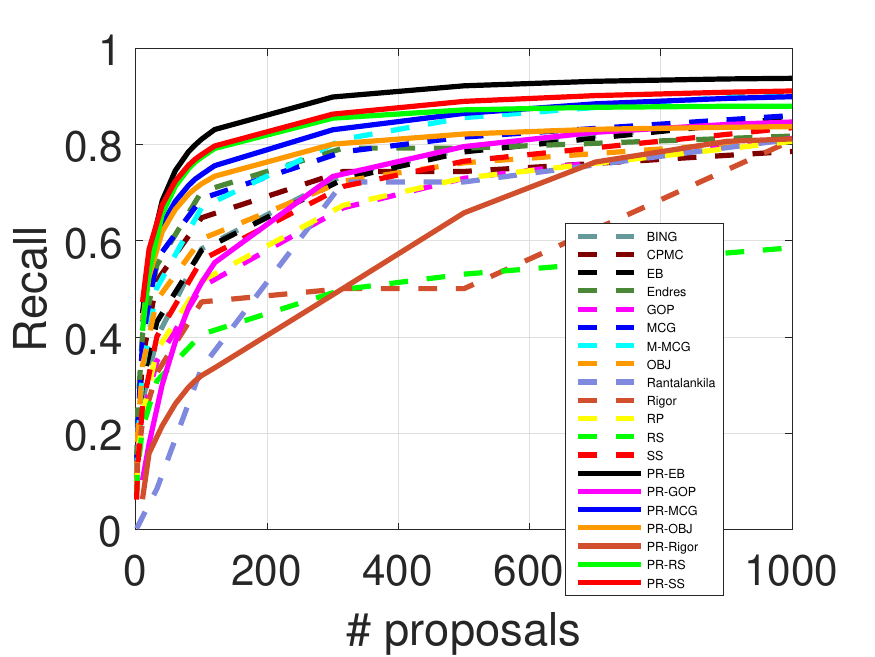}}

         \subfloat[Recall at 0.7 IoU.]{
  \includegraphics[width=0.5\textwidth]{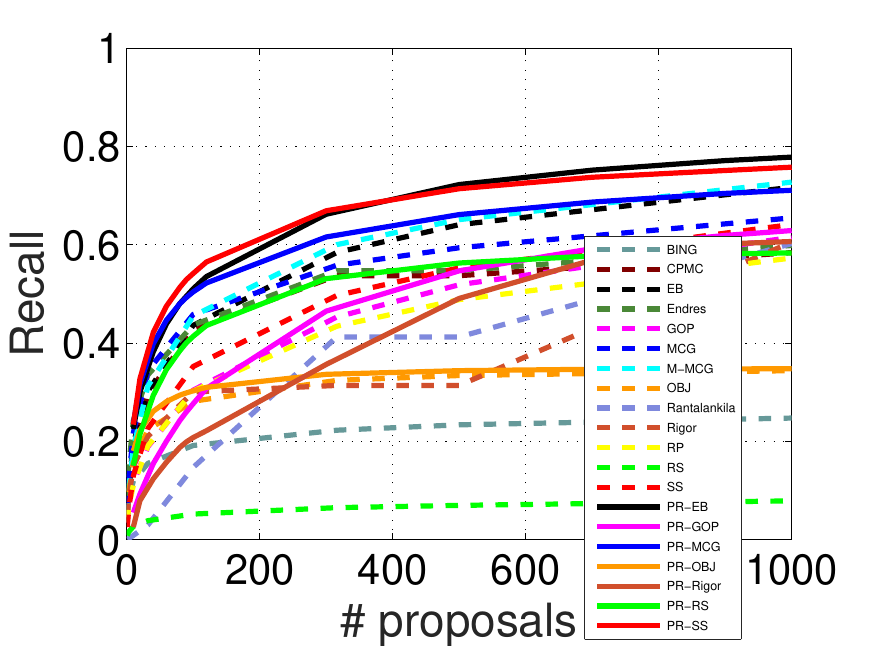}}
       \subfloat[Average recall.]{
  \includegraphics[width=0.5\textwidth]{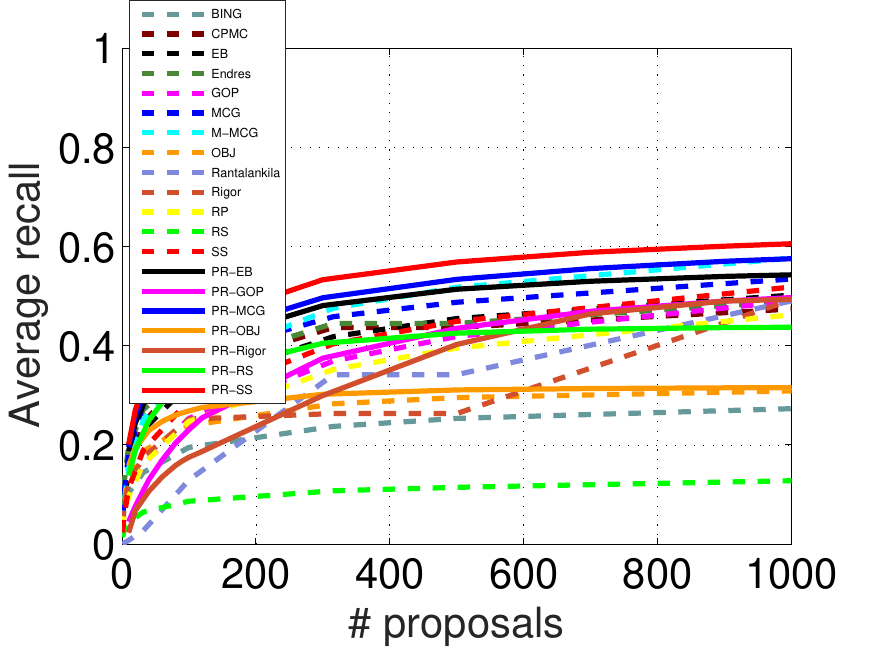}}

  \caption{\textbf{Comparison results with all baselines in terms of Recall versus IoU threshold (a), (b) and (c) and Recall versus number of proposals (d), (e) and (f).}}
\label{fig:methods_proposal}
\end{figure}

\begin{table}[h!]
\centering
\caption{Recall, Average Recall, Mean Average Best Overlaps and time cost on VOC 2007 test set with 300 proposals per image.}
\label{tab:prop_300}
	\resizebox{\columnwidth}{!}{ 	
\begin{tabular}{|l|l|l|l|l|l|r|r|r|r|}
\hline
\multirow{2}{*}{Methods} & \multicolumn{6}{c|}{Recall}                                  & \multicolumn{1}{c|}{AR} & \multirow{2}{*}{MABO} & \multicolumn{1}{c|}{\multirow{2}{*}{Time(s)}} \\ \cline{2-8}
                         & 0.5  & 0.6  & 0.7  & 0.8  & 0.9   & \multicolumn{1}{l|}{1.0} & [0.5,1.0]               &                       & \multicolumn{1}{c|}{}                         \\ \hline
BING                     & 0.73 & 0.50 & 0.22 & 0.07 & 0.02  & 0                   & 0.23                    & 0.56                  & 0.06                                          \\ \hline
CPMC                     & 0.74 & 0.64 & 0.53 & 0.39 & 0.22  & 0.001                   & 0.43                    & 0.66                  & 250                                           \\ \hline
EB                       & 0.72 & 0.67 & 0.58 & 0.37 & 0.10  & 0                   & 0.41                    & 0.64                  & 0.3                                           \\ \hline
Endres                   & 0.79 & 0.68 & 0.54 & 0.38 & 0.20  & 0.001                   & 0.35                    & 0.60                  & 100                                           \\ \hline
GOP                      & 0.66 & 0.57 & 0.45 & 0.31 & 0.16  & 0.001                   & 0.36                    & 0.60                  & 1.2                                           \\ \hline
MCG                      & 0.78 & 0.68 & 0.55 & 0.41 & 0.24  & \textbf{0.002}                   & 0.45                    & 0.69                  & 30                                            \\ \hline
M-MCG                    & 0.81 & 0.72 & 0.60 & 0.44 & 0.21  & 0                  & 0.47                    & 0.70                  & 30.2                                          \\ \hline
OBJ                      & 0.72 & 0.61 & 0.32 & 0.09 & 0.01  & 0                        & 0.28                    & 0.57                  & 3                                             \\ \hline
Rantalankila             & 0.72 & 0.56 & 0.41 & 0.26 & 0.11  & 0                   & 0.34                    & 0.62                  & 10                                            \\ \hline
Rigor                       &0.50	&0.41	&0.31	&0.22	&0.12	&0            & 0.26                    & 0.50                  & 10                                             \\ \hline
RP                       & 0.67 & 0.56 & 0.43 & 0.29 & 0.13  & 0.001                   & 0.35                    & 0.60                  & 1                                             \\ \hline

RS                       & 0.49 & 0.23 & 0.06 & 0.01 & 0.00  & 0                        & 0.10                    & 0.48                  & \textbf{0.001}                                          \\ \hline
SS                       & 0.71 & 0.61 & 0.49 & 0.35 & 0.19  & \textbf{0.002}                   & 0.40                    & 0.64                  & 10                                            \\ \hline
VGG                      & 0.88 & 0.81 & 0.66 & 0.29 & 0.03  & 0                        & 0.44                    & 0.45                  & 0.15                                          \\ \hline
ZF                       & 0.85 & 0.76 & 0.58 & 0.24 & 0.02  & 0                        & 0.40                    & 0.41                  & 0.03                                          \\ \hline \hline

PR-EB                    & \textbf{0.90} & \textbf{0.82} & 0.66 & 0.38 & 0.10  & 0                        & 0.48                    & 0.69                  & 0.32                                          \\ \hline
PR-GOP                   &0.74   &0.62  &0.47  &0.31  &0.14 &0            & 0.38                     & 0.64                 & 1.22                                         \\ \hline

PR-MCG                   & 0.83 & 0.74 & 0.62 & 0.46 & 0.25  & 0.001                        & 0.5                     & 0.70                  & 30.02                                         \\ \hline
PR-OBJ                   & 0.80 & 0.67 & 0.34 & 0.10 & 0.02  & 0                        & 0.3                     & 0.61                  & 3.02                                          \\ \hline

PR-Rigor      &0.50    &0.44    &0.37    &0.29   &0.19    &0.002                        & 0.31                    &  0.50                 & 10.02                                          \\ \hline

PR-RS                    & 0.86 & 0.73 & 0.53 & 0.28 & 0.07  & 0                        & 0.41                    & 0.69                  & 0.02                                          \\ \hline

PR-SS                    & 0.86 & 0.79 & \textbf{0.67} & \textbf{0.50} & \textbf{0.27}  & \textbf{0.002}                        & \textbf{0.53}                    & \textbf{0.73}                  & 10.02                                         \\ \hline
\end{tabular}}
\end{table}

\begin{table}[h!]
\centering
\caption{Average Recall on each 20 class of VOC 2007 test set with 300 proposals per image}
\label{tb:ar300}
	\resizebox{\columnwidth}{!}{ 	 	
\begin{tabular}{|l|c|c|c|c|c|c|c|c|l|l|l|l|l|l|l|l|l|l|l|l|}
\hline
\multicolumn{1}{|c|}{Algorithms} & aero      & \multicolumn{1}{l|}{bicycle} & \multicolumn{1}{l|}{bird} & \multicolumn{1}{l|}{boat} & bottle    & \multicolumn{1}{l|}{bus} & \multicolumn{1}{l|}{car} & \multicolumn{1}{l|}{cat} & chair & cow & table & dog & horse & mbike & person & plant & sheep & sofa & train & tv \\ \hline
BING  &0.27 &0.27 &0.22 &0.16 &0.13 &0.27 &0.20 &0.40 &0.19 &0.22 &0.32 &0.34 &0.28 &0.27 &0.24 &0.20 &0.21 &0.37 &0.34 &0.24 \\ \hline

CPMC  &0.55 &0.44 &0.44 &0.33 &0.17 &0.61 &0.43 &0.76 &0.34 &0.52 &0.56 &0.71 &0.57 &0.51 &0.37 &0.31 &0.47 &0.71 &0.62 &0.53 \\ \hline

EB &0.51 &0.51 &0.45 &0.37 &0.20 &0.58 &0.41 &0.61 &0.32 &0.50 &0.50 &0.62 &0.57 &0.52 &0.37 &0.32 &0.47 &0.57 &0.57 &0.55  \\ \hline

Endres  &0.49 &0.53 &0.40 &0.33 &0.18 &0.59 &0.47 &0.74 &0.39 &0.49 &0.60 &0.70 &0.56 &0.55 &0.37 &0.33 &0.46 &0.75 &0.66 &0.49 \\ \hline

GOP  &0.41 &0.43 &0.34 &0.26 &0.12 &0.55 &0.38 &0.72 &0.27 &0.37 &0.58 &0.63 &0.50 &0.47 &0.32 &0.26 &0.33 &0.67 &0.59 &0.39 \\ \hline

MCG  &0.52 &0.49 &0.42 &0.32 &0.25 &0.63 &0.45 &0.73 &0.39 &0.52 &0.53 &0.67 &0.57 &0.52 &0.42 &0.32 &0.47 &0.70 &0.63 &0.59  \\ \hline

M-MCG  &0.60 &0.52 &0.47 &0.39 &0.25 &0.62 &0.46 &0.73 &0.41 &0.55 &0.57 &0.70 &0.58 &0.55 &0.43 &0.35 &0.52 &0.71 &0.65 &0.58 \\ \hline
OBJ  &0.33 &0.33 &0.27 &0.23 &0.13 &0.41 &0.28 &0.45 &0.21 &0.29 &0.42 &0.41 &0.37 &0.31 &0.26 &0.21 &0.26 &0.46 &0.41 &0.30 \\ \hline

Rantalankila &0.45 &0.35 &0.35 &0.25 &0.16 &0.42 &0.35 &0.63 &0.33 &0.42 &0.44 &0.59 &0.40 &0.41 &0.25 &0.25 &0.35 &0.59 &0.44 &0.51 \\ \hline

Rigor &0.37 &0.29 &0.27 &0.22 &0.09 &0.40 &0.29 &0.62 &0.17 &0.30 &0.35 &0.52 &0.37 &0.36 &0.20 &0.16 &0.24 &0.55 &0.41 &0.31 \\ \hline

RP &0.54 &0.38 &0.33 &0.28 &0.13 &0.50 &0.34 &0.66 &0.31 &0.38 &0.55 &0.59 &0.43 &0.42 &0.28 &0.24 &0.37 &0.67 &0.52 &0.48 \\ \hline

RS &0.04 &0.14 &0.12 &0.08 &0.02 &0.13 &0.08 &0.19 &0.12 &0.13 &0.08 &0.18 &0.16 &0.16 &0.09 &0.13 &0.14 &0.10 &0.14 &0.19  \\ \hline

SS   &0.62 &0.48 &0.41 &0.33 &0.14 &0.56 &0.39 &0.72 &0.33 &0.44 &0.60 &0.68 &0.50 &0.49 &0.34 &0.27 &0.41 &0.71 &0.61 &0.50  \\ \hline

VGG  &0.42 &0.48 &0.41 &0.34 &\textbf{0.29} &0.44 &0.45 &0.54 &0.34 &0.50 &0.47 &0.55 &0.52 &0.49 &0.48 &0.36 &0.45 &0.52 &0.50 &0.47 \\ \hline

ZF   &0.40 &0.46 &0.36 &0.32 &0.23 &0.41 &0.41 &0.51 &0.27 &0.45 &0.47 &0.53 &0.49 &0.44 &0.43 &0.32 &0.40 &0.50 &0.48 &0.38 \\ \hline \hline

PR-EB  &0.51 &0.55 &0.49 &0.41 &0.33 &0.59 &0.48 &0.62 &0.42 &0.53 &0.52 &0.63 &0.57 &0.57 &0.45 &0.43 &0.53 &0.58 &0.56 &0.52 \\ \hline

PR-GOP  &0.42	&0.43	&0.38	&0.30	&0.17	&0.54	&0.43	&0.63	&0.33	&0.48	&0.50	&0.57	&0.43	&0.42	&0.33	&0.27	&0.42	&0.58	&0.50	&0.48 \\ \hline

PR-MCG &0.56 &0.55 &0.45 &0.38 &{0.28} &0.64 &0.50 &0.72 &0.41 &0.58 &0.61 &0.68 &0.59 &0.57 &0.46 &0.40 &0.52 &0.71 &0.66 &0.56 \\ \hline

PR-OBJ &0.34 &0.34 &0.29 &0.25 &0.17 &0.41 &0.29 &0.45 &0.24 &0.30 &0.44 &0.42 &0.38 &0.34 &0.28 &0.25 &0.27 &0.47 &0.41 &0.30 \\ \hline

PR-Rigor &0.50 &0.38 &0.32 &0.30 &0.09 &0.47 &0.34 &0.68 &0.21 &0.35 &0.48 &0.60 &0.47 &0.43 &0.22 &0.19 &0.29 &0.62 &0.56 &0.28 \\ \hline

PR-RS  &0.50 &0.45 &0.41 &0.35 &0.20 &0.53 &0.44 &0.59 &0.39 &0.46 &0.45 &0.55 &0.44 &0.46 &0.35 &0.36 &0.45 &0.56 &0.46 &0.53 \\ \hline

PR-SS  &\textbf{0.69} &\textbf{0.60} &\textbf{0.54} &\textbf{0.45} &0.27 &\textbf{0.66} &\textbf{0.52} &\textbf{0.79} &\textbf{0.51} &\textbf{0.57} &\textbf{0.70} &\textbf{0.76} &\textbf{0.61  }&\textbf{0.61} &\textbf{0.47} &\textbf{0.44} &\textbf{0.55} &\textbf{0.80} &\textbf{0.68} &\textbf{0.65} \\ \hline
\end{tabular}}
\end{table}
In this section, we thoroughly compare with all the baselines. First, we plot the curve of recall versus IoU threshold in Figure~\ref{fig:methods_proposal}(a), Figure~\ref{fig:methods_proposal}(b) and Figure~\ref{fig:methods_proposal}(c). Generally speaking, when equipped with the partial ranking model, the proposal generation methods consistently outperform their original versions. For instance, in case of 100 proposals per image with IoU threshold 0.5, the recall of SS is 56.04\%, while PR-SS gains about 20\% (78.40\%). In case of 500 proposals per image with IoU threshold 0.7, the recall of RS is only 6.96\%, while PR-RS reaches 56.30\%. Compared with other baselines, the methods equipped with our partial ranking model also favors a great advantage. For example, in case of 500 proposals per image with IoU threshold 0.5, the best recall of our model is achieved by PR-EB which reaches 92.23\%, while the best of the baselines is 85.62\% (M-MCG). Then, we plot the curve of recall and average recall v.s. number of proposals in Figure~\ref{fig:methods_proposal}(d), Figure~\ref{fig:methods_proposal}(e) and Figure~\ref{fig:methods_proposal}(f). In terms of IoU threshold 0.5, PR-EB favors the best recall. For example, with 300 proposals per image, PR-EB reaches the recall as 89.91\% while the best of the baseline methods only reaches 81.14\% (M-MCG). As shown in Figure~\ref{fig:methods_proposal}b, with 500 proposals per image, the recall of PR-SS is the best with 54.52\%, while SS is 40.27\%. The partial ranking model again dramatically improves the original model. Note that RS is regular sampling process and inferior than all the others in Figure \ref{fig:methods_proposal}(a). But our partial ranking model guarantees a great advantage over RS. That is, PR-RS reaches similar recall as PR-SS and only inferior than PR-EB. In terms of average recall, PR-SS achieves the best result over all the comparative algorithms as shown in Figure \ref{fig:methods_proposal}(c). Specifically, in case of 500 proposals per image, the average recall of PR-RS is (0.43),  better than BING (0.25), CPMC (0.43), Endress (0.44), GOP (0.41), OBJ (0.29), Rantalankila (0.34), Rigor (0.26), RP(0.39), RS(0.11), and as comparative as EB (0.45) and SS(0.44). PR-SS achieves the best result with 0.59. 

Then we report the recall, average recall, mean average best overlaps and the time cost in Table~\ref{tab:prop_300}. The time cost of baseline methods are provided by \cite{Hosang2015pami} and the related paper \cite{ren2015faster}. Though VGG and ZF use deep convolution network for proposal generation, our improved model PR-SS obtains the best recall in most cases, the best average recall and mean average best overlaps. VGG and ZF is efficient due to the use of GPU computation, while others are computed on CPU. But our partial ranking process is still efficient with time cost less about 0.02 second. It consists of HOG feature extraction (0.0035 seconds), CWS sampling (0.019 seconds) and partial ranking (2.69e-4 seconds). The code is run on Matlab 2015 with Windows 8 system, CPU 3.10GHz. In summary, our model enjoys a balance between time efficiency and recall.

Finally, we record the average recall on each class of VOC 2007 test set in Table~\ref{tb:ar300}. In most cases, we find that our partial ranking model gains a great advantage over the baselines. This agrees with our previous observation that min-max kernel with partial ranking always leads to appealing performance. That is, PR-SS achieves the best average recall on 19 over 20 classes. Except on the class ``bottle'', though VGG obtains the best average recall with 0.29, PR-SS obtains comparative result as 0.27. As shown in \cite{Hosang2015pami}, average recall is closely correlated with detection performance. Hence, the obtained results indicate the proposals produced by our algorithms are favourable for further vision tasks.

\section{Conclusion}
\label{sec:conclu}
In this paper, based on the observation that it is typically not necessary to derive a full ranking for the total candidates, we propose a new partial ranking model for object proposal. The main difference of our model and other full ranking models, such as \svmrank, is that we only constrain the relative orders of the two subsets: the top-$k$ candidates and the last $n-k$ candidates. We then show that such a model can be equivalently transformed into the large margin based framework, in the sense of keeping the relative ranks of the two subsets. Furthermore, we demonstrate that our model can easily be kernelized, and the broadly used resemblance kernel can be approximated by linear function which is with great efficiency. In the experiments, we show that the non-linear kernel and partial ranking model always result in improved performance in terms of recall and average recall. Our future works include speeding up our partial ranking process and integrating with the deep networks for specific vision tasks.

\section*{References}

\bibliography{obj}

\end{document}